\documentclass[conference]{IEEEtran}
\usepackage[T1]{fontenc}
\usepackage{graphicx,booktabs,array,tabularx,float,placeins,url}
\usepackage[hidelinks]{hyperref}
\usepackage[expansion=false]{microtype}
\usepackage{xurl}
\usepackage{amsmath}
\newcolumntype{Y}{>{\raggedright\arraybackslash}X}
\usepackage{etoolbox}
\patchcmd{\abstract}{\textit{\abstractname}---}{}{}{\PackageError{manuscript}{Could not suppress abstract label}{}}
\patchcmd{\IEEEkeywords}{\textit{\IEEEkeywordsname}---}{\textit{\IEEEkeywordsname}: }{}{\PackageError{manuscript}{Could not replace keyword separator}{}}
\begin{document}
\raggedbottom
\setcounter{topnumber}{3}
\setcounter{dbltopnumber}{2}
\renewcommand{\topfraction}{.9}
\renewcommand{\dbltopfraction}{.9}
\renewcommand{\textfraction}{.08}
\renewcommand{\floatpagefraction}{.75}
\renewcommand{\dblfloatpagefraction}{.75}
\author{\begin{minipage}{\textwidth}\centering\normalsize
Okenla AbdulQahhar Abimbola\textsuperscript{1,*},
Maurice Abraham John\textsuperscript{1},
Adepoju Jeremiah Moyondafoluwa\textsuperscript{1},
Omolaiye Samuel\textsuperscript{1},
Oko-Oba Sandra Chioma\textsuperscript{1},
Owoo Oluwamayowa Emmanuel\textsuperscript{1},
Thanni Adewuyi\textsuperscript{1},
Lily Aimas Love\textsuperscript{1},
Benson Eniola\textsuperscript{1},
Abdulquddus Ajibade\textsuperscript{1},
Sylvia Essessien\textsuperscript{1},
Anyanwu Princess Emmanuella\textsuperscript{1},
Adereni Abiodun\textsuperscript{1}\par\vspace{0.8ex}
\textsuperscript{1}HelpMum Africa\par\vspace{0.5ex}
Corresponding author: Okenla AbdulQahhar Abimbola\par
Email: \href{mailto:aitech@helpmum.org}{aitech@helpmum.org}
\end{minipage}}
\title{Offline-First Quantized AI: Design and Evaluation of an On-Device Maternal-Health Assistant for Low-Connectivity Environments}
\maketitle

\begin{abstract}
Offline language models could extend access to maternal-health information where internet connectivity is unreliable, but deployment requires both clinically appropriate responses and feasible local execution. This study evaluated an Android assistant using ready-made quantized models, a clinically validated local knowledge package, conversation history, and optional local retrieval. Gemma 4 E2B and Qwen3.5 0.8B were compared on 45 conversations and 312 turns per configuration. Gemma retained context more consistently, whereas Qwen generated faster and used less memory. Clinician assessment of 150 Gemma responses across nine clinical domains yielded a mean quality score of 3.80/4, with 147 responses meeting the acceptance criterion. In 74 matched RAG-on/off questions, 11 RAG-on answers improved, 52 were comparable, four were degraded by returned content, and seven were lower-rated without clear evidence that retrieval caused the difference. Full-cohort mean differences were small and uncertain. In a sensitivity analysis excluding flagged cases, RAG-on had descriptively higher mean usefulness, safety, correctness, and actionability in the conditional 70-pair subset excluding the four flagged pairs from both conditions. Physical-device testing demonstrated offline responses on a 6 GB Android handset, with a 14.3 s median latency, alongside successful reconnection synchronization in the integrated application. Together, the findings characterize the integrated offline maternal-health assistant within the designed and tested benchmark, retrieval pipeline, and handset conditions.
\end{abstract}
\section{Introduction}
Nigeria's maternal health burden makes timely, understandable information a consequential design problem. The United Nations inter-agency estimates for 2023 attribute approximately 75,000 maternal deaths, or 28.7\% of the global total, to Nigeria; the estimated maternal mortality ratio was 993 deaths per 100,000 live births, with an uncertainty interval of 718--1540 \cite{r22}. This burden motivates accessible educational support that can explain routine care and distinguish warning signs requiring prompt assessment, consistent with antenatal and postnatal care recommendations \cite{r16,r17}. An assistant that explains routine concerns adequately but weakens the urgency of a warning-sign response would be poorly suited to this setting.

Access to a mobile network does not guarantee continuous access to a remote AI service. The Nigerian Communications Commission's Q3 2025 measurements document persistent urban--rural differences in mobile broadband experience, including differences where network coverage is present \cite{r23}. At the global level, the International Telecommunication Union estimated that 2.2 billion people remained offline in 2025 and identified continuing affordability and service-quality gaps \cite{r24}. These sources motivate a specific availability requirement: after installation of its model and knowledge package, an assistant should be able to answer supported questions without a network round trip. The system design consequently treats network-independent access as an explicit requirement \cite{r18}.

Prior work addresses different parts of this problem. ADVISER documents how an AI-supported vaccination intervention operated in Nigeria with health workers and delivery constraints \cite{r25}. Ajibade et al. compare maternal-health and vaccination model responses after domain adaptation \cite{r1}. MamaBench tests whether a model changes its answer when a clinically decisive detail changes in a maternal or paediatric presentation \cite{r26}. None of these results alone answers whether the present assistant retains context over a conversation, gives the required care-seeking advice, and runs locally on the tested phone. Those questions determine the experiments here.

Compact local inference introduces an additional trade-off. Reducing the deployment footprint can make execution feasible on a mobile device, yet a fast response may omit the information needed for the user's decision. Conversely, a configuration with better conversational behavior may require more usable memory than the device can sustain. MobileLLM addresses efficient architectures for on-device tasks \cite{r13}, while measured resource and response trade-offs remain properties of the adopted application configuration. For a maternal health application, the relevant selection problem is the behavior of the complete adopted configuration: model artifact, native runtime, prompt policy, retained context, and retrieval. Because the intended system is conversational, continuity across turns is an upstream requirement rather than an optional quality dimension: a model must carry forward facts, respond to newly introduced urgency, and preserve the requested action as the exchange evolves. Conversational diagnostic evaluation treats these dialogue capabilities as distinct from isolated answer quality \cite{r10}, and capability-specific behavioral testing provides a basis for screening them separately \cite{r27}. A dedicated multi-turn action benchmark additionally probes the model's capacity to operate within an offline agentic interface.

Retrieval adds another dependency. Clinically validated content supplies a stronger basis for an answer only when the returned passage fits the current question and the model uses it appropriately. Medical RAG evaluations have shown that the corpus, retriever, and generator interact \cite{r3}. An accepted passage can still be incomplete, refer to the wrong life stage, or answer a different practical request. Maternal chatbot research has therefore combined response evaluation with triage and retrieval assessment \cite{r2,r7}. Offline operation makes this issue especially important because a local package cannot rely on unrestricted remote search to repair a poor match at answer time. Making retrieval a model-invoked local tool allows evidence acquisition to respond to the conversational information need, following the broader move toward adaptive and agentic RAG \cite{r50,r53}.

This study asks how well a compact Android assistant maintains maternal-health context, answers reviewed questions, uses local evidence, and responds offline. Its contribution is a linked evaluation of model choice, clinical response, retrieval effects, and local execution: conversation and action scores for two candidates, clinician ratings for the selected model, an audit of matched responses with and without passages, and handset timing. The sequence matters because Qwen's faster execution accompanied weaker context retention, while Gemma's strong isolated clinical scores did not guarantee that each retrieved passage would improve an answer.

\section{Related Work}
\subsection{Maternal health conversational systems and deployment}
Amil et al. review pregnancy and postpartum conversational agents whose evaluations span user experience and health-related responses \cite{r21}. Those measures capture different parts of use: a usable interface may still give incomplete warning-sign advice, while a sound answer to a fixed question says little about daily use. The experiments here examine answer quality and local execution.

HerCare grounds women's-health answers in two source collections \cite{r2}; Jha et al. combine pregnancy-stage routing with retrieval and expert assessment in India \cite{r7}; and Ajibade et al. report different effects of domain adaptation on maternal-health and vaccination answers in Nigeria \cite{r1}. Their results motivate a linked test of sustained conversation, the effect of individual retrieved passages, and Android CPU execution for the reviewed local package.

ADVISER's Nigerian implementation served more than 13,000 families and documented changes to costs, data collection, and intervention delivery \cite{r25}. It illustrates why an intervention's operating conditions deserve measurement alongside the information it provides \cite{r19}.

\subsection{Clinical language model evaluation}
Medical knowledge and conversational behavior require different evaluation designs. Singhal et al. assess clinical knowledge \cite{r9}, while Tu et al. investigate diagnostic dialogue and its communication demands \cite{r10}. HealthBench uses physician-authored criteria for open-ended health conversations \cite{r4}, and CLEVER formalizes expert review and reliability assessment \cite{r5}. These approaches support explicit clinical keys and case-sensitive evaluation rather than a single generic impression of answer quality.

MamaBench tests sensitivity to clinically meaningful differences using 217 counterfactual pairs of expert-authored maternal and paediatric narratives \cite{r26}. Its relevance to the present study is the requirement that a response reflect the specific presentation, including the details that change the appropriate action. Here, that principle informs assessment of changing urgency across conversations and of whether retrieved guidance preserves the action required by the full question \cite{r26,r27}. The evaluation focuses on educational guidance and care-seeking advice within the assistant's intended scope.

Behavioral testing in NLP provides a broader basis for examining such dependencies. CheckList organizes tests around capabilities and failure-revealing cases \cite{r27}. LLM judging makes open-ended review more scalable, but documented position and verbosity biases require attention \cite{r11}. The evaluation distinguishes factual correctness, responsiveness to the question, practical usefulness, and actionability. A longer response does not earn additional credit merely by containing more text, and a shorter answer can be sufficient when it gives the required explanation or decision.

\subsection{Retrieval and compact local inference}
RAG combines external evidence with language generation at inference time \cite{r12}. The MIRAGE and MedRAG study evaluates medical questions across alternative retrieval configurations, demonstrating the need to assess component choices rather than assuming a universal RAG benefit \cite{r3}. Research on irrelevant context also shows that additional input can distract a model; that evidence comes from a different task and supplies a possible mechanism rather than proof of the cause of the observed clinical failures \cite{r28}.

Pocket RAG studies offline first-aid guidance on Android \cite{r8}, while MobileLLM investigates small models designed for local execution \cite{r13}. The maternal-health workload brings these concerns together: its selected model must sustain conversation, fit the device, and use returned clinical passages appropriately. The comparison measures those behaviors for two distributed artifacts.

PalmBench evaluates compressed language-model configurations on mobile platforms, reinforcing the joint assessment of quality and execution cost \cite{r33}. MAM-AI investigates an offline medical retrieval system for nurses and midwives in Zanzibar using on-device embedding and guideline retrieval \cite{r34}. These systems supply related deployment precedents; the present evaluation focuses on a clinically validated question--answer package, maternal-health educational needs, conversation context, and action tasks used specifically to benchmark offline agentic capability.

Post-training quantization research provides the background for compact deployment: GPTQ studies weight reconstruction using approximate second-order information, while AWQ uses activation-aware scaling \cite{r39,r40}. The distributed artifacts evaluated here retain their original precision formats and native runtimes.

MELT measures runtime, memory, and energy on mobile and edge devices \cite{r37}; MobileAIBench examines both task results and resource use on physical devices \cite{r38}. They show why a quality score alone cannot establish mobile feasibility. HerCare, Jha et al., and MAM-AI examine health-domain retrieval and responses \cite{r2,r7,r34}, but provide different device and clinical evidence. This study reports the two kinds of evidence for one application: clinical and conversational ratings, paired answer changes when passages are supplied, emulator comparisons, and a small handset execution test. Each result has its own workload and denominator.

\subsection{Agentic retrieval and conversational evidence use}
Agentic RAG places retrieval within a model-directed interaction loop rather than treating evidence acquisition as an obligatory preprocessing step. Treating retrieval as an optional action allows the response process to alternate between answering from available context and seeking additional evidence \cite{r49,r53}. Self-RAG learns retrieval and reflection decisions through additional training and special tokens \cite{r50}. Adaptive-RAG uses a trained question-complexity classifier to choose among no retrieval, single-step retrieval, and iterative retrieval \cite{r51}. These approaches establish distinct ways to condition evidence acquisition on the information need; they do not imply that every adaptive system requires the same training procedure or retrieval machinery.

In this application, the existing local model can request the local retrieval tool. A deterministic search then returns passages that pass admission checks. This uses the model's existing function-calling interface; the model was not retrained. Ni et al. show that a model's confidence can be a poor guide to whether it needs retrieval \cite{r52}. The relevant questions here are therefore whether the model requests evidence, whether the returned text fits the request, and whether the final answer uses it correctly. The paired RAG-on/off experiment tests the last of these questions by supplying accepted passages under controlled conditions; it does not measure the model's live decision to call the tool.

\section{Methods}
The methods follow the path from clinical requirements to observed answers. The safety boundary and question bank define what a suitable response must contain. The local knowledge and system sections describe what information the model can use. Candidate comparison and the action benchmark explain model selection; the scoring and analysis sections define how clinical answers, conversations, retrieval effects, and device execution were measured.

\subsection{Study design and intended use}
The system is intended to provide maternal, newborn, and vaccination information to pregnant and postpartum people, caregivers, and community health workers. Its educational scope includes routine pregnancy questions, antenatal preparation, nutrition, common concerns, medication and exposure information, postpartum care, newborn care, and vaccination. WHO antenatal and postnatal recommendations provide clinical context for these domains \cite{r16,r17}. The benchmark also includes uncertainty, practical access barriers, warning signs, and requests that the assistant should decline or refer.

At each turn, the application passes the current question and bounded conversation history to the local model. The model may answer directly or request passages from the local package. Isolated clinical responses are scored against the question's expected facts, requested action, and critical-error conditions in the separate clinician-reviewed bank. The multi-turn workload is limited to context retention and a scripted action benchmark that tests whether the model selects a simulated offline action, carries facts across turns, supplies valid arguments, and reports the execution result accurately. This separation prevents conversational continuity and tool behavior from being interpreted as clinical safety ratings.

The application exposes model-initiated local retrieval as its only deployed tool. The multi-turn comparison uses a separate, dedicated action interface to benchmark offline agentic capability; its tools are experimental components and are not application features being deployed. The paired retrieval experiment is a controlled evidence-context comparison rather than a tool-use benchmark: accepted passages were supplied to the RAG-on generation and omitted from RAG-off, with the model-facing retrieval tool disabled in both arms. The paired count therefore measures retrieval-eligible evidence contexts, not the model's decision to invoke retrieval.

\subsection{Evaluation sequence and analysis units}
The clinical and conversational experiments use three evaluation units. A clinical question defines an isolated information need and its expected facts and action. A conversation tests behavior across ordered turns, including whether a later answer uses or updates earlier context. A matched question pair compares the selected model with the retrieval configuration enabled and disabled. Execution events and process measurements are retained separately from clinical ratings. A supplementary device-envelope assessment examines local generation, recorded response latency, and reconnection synchronization in the integrated application.

Candidate selection preceded clinical response validation. Both configurations first completed the common isolated-question, multi-turn, action, and execution workloads. The multi-turn results were used as the system-level screen because the intended assistant must preserve context before an isolated answer can be considered sufficient. The configuration with the stronger conversational profile was then carried forward for clinician-reviewed single-question responses and the paired retrieval experiment. This staged sequence keeps model screening, clinical response validation, and retrieval analysis distinct, consistent with staged reporting of AI systems from technical evaluation toward clinical assessment \cite{r6,r20}.

The experiments address three research questions. RQ1 asks how the candidate configurations trade context retention and benchmark action behavior against latency and memory. RQ2 asks how accepted retrieval changes usefulness, safety, correctness, and actionability for the selected configuration. RQ3 asks which retrieval-linked and conversational behavior patterns remain despite successful execution and a clinically validated knowledge package. These questions connect the clinical response requirements to the practical operating conditions of the system \cite{r6,r19}.

\subsection{Clinical Safety Boundary}
The clinical safety boundary determines the response priority before optional explanation or any action required by a benchmark scenario. Four levels distinguish routine education from uncertainty, potential warning signs, and emergency presentations (Table~\ref{tab:1}). These levels are assigned to benchmark scenarios during clinical design and review; they are not labels inferred retrospectively from model performance. The response is assessed against the action required by the complete scenario, including pregnancy stage, symptom pattern, and urgency. This approach translates antenatal and postnatal recommendations into observable response requirements \cite{r16,r17}.

\begin{table}[!tbp]
\centering
\caption{Clinical safety boundary}
\label{tab:1}
\small
\setlength{\tabcolsep}{3pt}
\renewcommand{\arraystretch}{1.16}
\begin{tabular}{@{}>{\raggedright\arraybackslash}p{0.13\linewidth} >{\raggedright\arraybackslash}p{0.34\linewidth} >{\raggedright\arraybackslash}p{0.45\linewidth}@{}}
\toprule
\textbf{Level} & \textbf{Scenario} & \textbf{Required response} \\
\midrule
E0 & Routine education without an apparent warning sign. & Answer the supported question clearly and provide relevant clinical guidance. \\
E1 & Unclear, unsupported, incomplete, or conflicting information. & State the boundary; seek the minimum useful clarification and recommend care when appropriate. \\
E2 & Potential warning sign with uncertain severity or urgency. & Prioritize safety and prompt assessment; limit clarification to details necessary for the action. \\
E3 & Emergency concern, including severe bleeding, seizure, loss of consciousness, or severe breathing difficulty. & Advise immediate in-person emergency care. Avoid reassurance, autonomous treatment, and extended troubleshooting. \\
\bottomrule
\end{tabular}
\end{table}

Clinical response requirements remain distinct from passage retrieval. A passage about the same broad topic cannot downgrade an emergency or replace the question-specific care recommendation. When escalation is required, the care instruction precedes optional background and requests for nonessential details. The assistant does not make autonomous diagnoses, prescribe individualized treatment, change medication, or invent service availability and contact numbers. Conversely, a routine educational question should receive the supported explanation rather than a generic referral. This boundary makes both inappropriate reassurance and unnecessary abstention assessable behaviors.

\subsection{Evaluation Datasets}
\label{sec:datasets}
The evaluation follows four linked cohorts. The clinical bank contains 150 questions; both candidates generated isolated answers, and the selected Gemma answers received clinician scoring. The conversation benchmark contains 45 conversations and 312 turns for each candidate. Retrieval yielded 74 matched Gemma questions with and without accepted passages. The 70-pair score panel is a secondary view of those same pairs after four flagged identifiers were set aside for case analysis. Table~\ref{tab:3} gives the unit and purpose of each workload.

In-house medical doctors developed and reviewed the 150-question bank for relevance and usefulness. Each item carries a stable identifier, clinical domain, persona, safety level, source mapping, expected facts, expected action, and critical-error conditions. The review process separates whether a question is appropriate for the intended application from how a generated response subsequently performs. The domain profile in Table~\ref{tab:2} covers routine education, contextual decisions, and situations requiring assessment or an explicit information boundary. The intended population includes pregnant and postpartum women, mothers, caregivers, and frontline health workers in African settings with constrained access to care and connectivity.

Forty items marked for revision in the consensus review record were revised using the existing clinical comments and reference mappings. The revisions made decision-relevant context explicit: gestational or postpartum timing, symptom severity, onset, prior assessment, medication exposure, missing records, or access constraints. Expected facts and action keys were updated to remain consistent with each revised scenario. Stable identifiers preserved the link between the question, collection record, and scoring worksheet.

The benchmark keys define acceptable behavior without requiring one exact wording. A response can explain uncertainty, give a practical step, or recommend assessment without reproducing every optional reference detail. Critical-error conditions address consequences such as missed escalation, inappropriate reassurance, unsupported treatment, and dishonest claims about a service or local action. The construction follows the case-specific evaluation principle used in clinical conversation benchmarks \cite{r4}.

\begin{table}[!tbp]
\centering
\caption{Clinical question bank domain profile}
\label{tab:2}
\small
\setlength{\tabcolsep}{3pt}
\renewcommand{\arraystretch}{1.16}
\begin{tabular}{@{}>{\raggedright\arraybackslash}p{0.27\linewidth} >{\raggedright\arraybackslash}p{0.1\linewidth} >{\raggedright\arraybackslash}p{0.55\linewidth}@{}}
\toprule
\textbf{Domain} & \textbf{Items} & \textbf{Clinical and communication focus} \\
\midrule
Pregnancy milestones & 15 & Early changes, pregnancy timing, normal variation, and antenatal engagement. \\
Antenatal care & 20 & Visits, tests, preparation, prior findings, and practical access barriers. \\
Nutrition and exercise & 15 & Diet, activity, pica, weakness, and context-sensitive advice. \\
Common concerns & 15 & Symptom explanation, severity, timing, and appropriate assessment. \\
Medication and exposure & 15 & Treatment boundaries, uncertain exposures, supplements, and unsupported claims. \\
Postpartum and newborn care & 20 & Recovery, feeding, newborn concerns, and escalation when needed. \\
Vaccination and immunization & 20 & Schedules, missing records, eligibility questions, and service information boundaries. \\
Warning signs and escalation & 20 & Recognition of consequential symptom patterns and clear care-seeking urgency. \\
Unsupported or review-required & 10 & Uncertainty, out-of-scope requests, and avoidance of invented information. \\
\bottomrule
\end{tabular}
\end{table}

The safety-level distribution is E0, 51 questions; E1, 46; E2, 33; and E3, 20. Domain and safety level describe different properties: a medication question can concern routine information or a consequential exposure, while a postpartum question can involve either ordinary recovery or an emergency. The question-bank profile therefore supports interpretation of performance across content and urgency without treating every item within a domain as equally hazardous.

Table~\ref{tab:3} distinguishes the clinical bank from the conversation, retrieval, and execution workloads. The same question identifiers link the clinical keys to preserved responses and scoring rows. The clinician response workbook records the prompt, expected facts, expected action, source mappings, generated answer, dimension ratings, critical-error flags, and acceptance outcome. These linked records maintain the relationship between the clinical request, assessment criteria, and aggregate findings \cite{r4,r5}.

\begin{table}[!tbp]
\centering
\caption{Evaluation datasets and their analytical roles}
\label{tab:3}
\small
\setlength{\tabcolsep}{3pt}
\renewcommand{\arraystretch}{1.16}
\begin{tabular}{@{}>{\raggedright\arraybackslash}p{0.2\linewidth} >{\raggedright\arraybackslash}p{0.2\linewidth} >{\raggedright\arraybackslash}p{0.52\linewidth}@{}}
\toprule
\textbf{Dataset} & \textbf{Composition} & \textbf{Purpose and unit of analysis} \\
\midrule
Clinical questions & 150 questions & Nine-domain bank with reviewed facts, actions, references, and critical-error conditions. Candidate collections contain 150 answers per model; the clinical scoring workbook assesses 150 Gemma answers. \\
Conversation workload & 45 conversations; 312 turns per model & Ordered exchanges assess retained context, changing urgency, and benchmark-only offline actions. Ratings use conversations; completion and execution errors use turns. \\
Paired retrieval cohort & 74 matched questions & Selected-model answers with retrieval enabled and disabled. The primary analysis retains all 74 pairs; a secondary sensitivity display includes 70 pairs, while the separate audit retains the four flagged cases. \\
Retrieval verification & 25 known-item queries & Checks passage recovery in the original normalized package. This component test is separate from clinical response scoring and from coverage in the revised active package. \\
Execution measurements & Workload-specific records & Android CPU emulator latency, throughput, memory, completion, and action errors remain linked to their workloads. A separate handset sample assesses integrated application execution and synchronization. \\
\bottomrule
\end{tabular}
\end{table}

\subsection{Benchmark governance and review roles}
In-house medical doctors developed and reviewed the question bank, its expected facts and actions, and critical-error conditions. System development provided the candidate artifacts, prompts, retrieval implementation, and preserved outputs. Stable question identifiers linked the outputs to the reviewed keys and scoring workbooks. Clinicians rated the 150 selected-model answers; the conversation and paired RAG comparisons used rubric-based LLM judgments. These are distinct scoring streams, with their units, criteria, and outcomes reported separately. Knowledge-package curation also took place separately from benchmark construction: benchmark questions and model outputs were not used to select or rewrite passages (Section~\ref{sec:knowledge}). These records make the origin of the criteria, responses, and ratings traceable in the internal evaluation \cite{r4,r6}.

\subsection{Local Knowledge Package and Retrieval}
\label{sec:knowledge}
A device-resident maternal-health knowledge package was constructed from first-party MamaBot AI and vaccination chat-log exports. The material was cleaned, checked against clinical sources, and clinically validated before inclusion in the active corpus. Table~\ref{tab:4} distinguishes the original export, normalized records, reviewed content, and the passages admitted to retrieval. These counts describe successive preparation stages rather than interchangeable estimates of corpus size.

\begin{table}[!tbp]
\centering
\caption{Local knowledge package composition and clinical provenance}
\label{tab:4}
\small
\setlength{\tabcolsep}{3pt}
\renewcommand{\arraystretch}{1.16}
\begin{tabular}{@{}>{\raggedright\arraybackslash}p{0.29\linewidth} >{\raggedright\arraybackslash}p{0.1\linewidth} >{\raggedright\arraybackslash}p{0.53\linewidth}@{}}
\toprule
\textbf{Preparation stage} & \textbf{Count} & \textbf{Role in the knowledge resource} \\
\midrule
Source export & 36,636 & Original maternal-health and vaccination conversation records. \\
Normalized question--answer records & 36,548 & Complete records formed during initial preparation; 88 source records remained unsplit. \\
Clinically reviewed content & 23,977 & Reviewed rows, including ten supplemental coverage records; original and final text remain linked through the audit. \\
Active retrieval passages & 21,673 & Clinically validated complete question--answer passages admitted to the current package. \\
Excluded reviewed rows & 2,304 & Retained outside the active corpus for audit and disposition review. \\
Clinical source register & 28 & Source entries covering Nigerian and international institutional guidance. \\
\bottomrule
\end{tabular}
\end{table}

The source material covers pregnancy and antenatal care, birth preparation, postpartum recovery, newborn care, breastfeeding, nutrition, and vaccination. Clinical review addressed warning signs, haemorrhage, hypertensive disorders, perinatal mental health, gestational diabetes, maternal HIV, anaemia, malaria, medication information, and service localization. A register of 28 source entries draws on Nigerian Federal Ministry of Health guidance, NPHCDA, UNICEF Nigeria, and WHO material. The review therefore considers both the clinical claim and whether its wording and recommended action fit the intended setting \cite{r16,r17}.

Each passage preserves a complete question-and-answer record. The audit links the original question and answer to the reviewed text, disposition, identified issues, severity, evidence references, and decision note. Exclusions retain an inspectable record of out-of-scope material, unsupported service claims, time-sensitive information, and content lacking an adequate clinical source. Where an information gap remains within scope, the boundary is stated explicitly rather than filled with an unsupported claim. The active package is versioned separately from mutable conversation and user-document records, allowing a content update to be reviewed and replaced without altering the stored conversation history.

Knowledge curation was conducted separately from benchmark construction. Benchmark questions, expected answers, retrieval-evaluation queries, and model outputs were not inputs for selecting or rewriting knowledge passages. This separation preserves the role of the package as a clinical information resource rather than an answer set constructed around the evaluation questions.

The current architecture separates the decision to seek evidence from the decision to admit a passage. The local model receives the current question, bounded conversation history, response policy, and access to a local retrieval tool. It can answer directly or request local evidence before continuing its answer. The runtime binds the tool to the current user message and user/session scope; the model supplies no query, identity, or access-control arguments. This implements selective evidence acquisition through a local tool, following the model-directed retrieval principle \cite{r49,r50}.

When invoked, the tool searches the on-device question and answer fields using SQLite FTS5/BM25 \cite{r29}. Vocabulary normalization covers supported clinical equivalents. Deterministic lexical, metadata, and requested-action checks assess topic, life stage, population, timing, and country scope. The underlying service assembles a pool of at most three candidates and applies the admission thresholds in Appendix B; the model-facing tool returns at most two accepted passages within the retrieved-text budget. It returns answer text, separated by newlines, through a native tool-result message. Passage identifiers and provenance remain available to the application separately. An empty result supplies no supporting evidence.

A high retrieval score does not establish that a passage answers the clinical question. A candidate can rank well against other available passages while still omitting the symptom or action that matters. The evaluation therefore retains retrieved text and judges its fit separately from the fact that it passed admission. This distinction is central to RQ2: retrieval availability and clinically appropriate evidence are distinct aspects of pipeline evaluation \cite{r3,r31}.

The agentic decision is when to request retrieval; it does not replace lexical search with model-based ranking or permit unrestricted database access. No additional embedding model, query-rewriting model, or learned reranker is installed. This preserves a small retrieval footprint and an inspectable admission path. Because the query remains the current message, conversation context guides invocation without automatically expanding an abbreviated follow-up into a standalone search query. The policy asks for a short clarification when a missing detail would materially change the guidance.

Retrieved text supports the current request and is excluded from later conversation history. Citation objects are retained separately from the generated answer, allowing evidence selection and answer content to be inspected independently. User-supplied documents form a separate source class, with user and session filters applied before results are returned. Uploaded material does not acquire the clinical authority of the validated package. The same local interface can therefore supply contextual documents while preserving the distinction between personal material and reviewed clinical guidance.

\subsection{System architecture and conversation context}
\begin{figure*}[!t]
\centering
\includegraphics[width=\textwidth,height=.58\textheight,keepaspectratio]{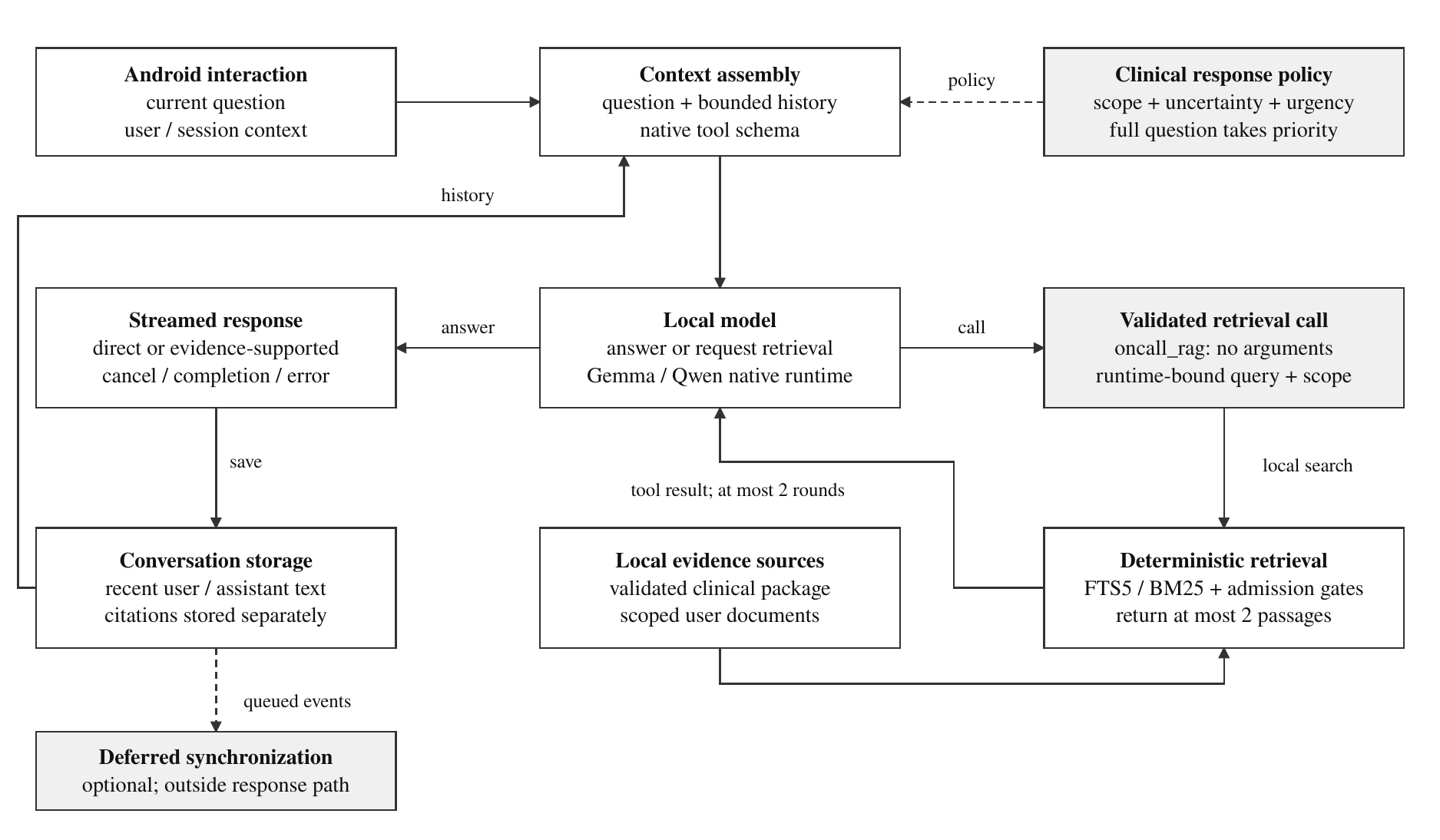}
\caption{Offline architecture with model-initiated retrieval. The generator answers directly or requests the allowlisted local tool. Deterministically admitted evidence returns through a bounded tool loop; conversation history, clinical response policy, and separately stored citation metadata retain distinct roles.}
\label{fig:1}
\end{figure*}

The assistant combines native local generation, clinically validated knowledge, bounded conversation history, and model-initiated retrieval (Figure~\ref{fig:1}). After installation, both a direct answer and an evidence-supported answer can be produced without a network service. The generator chooses whether to invoke retrieval; the application controls which tool can execute, which sources can be searched, and how much evidence returns. Conversation storage and optional synchronization remain outside the immutable clinical package.

The context builder keeps the current question distinct from supporting evidence. The first generation round receives the response policy and recent conversation, with retrieval available as a native tool. Evidence returned by a call informs the next round of the same request; it is not promoted into system policy or retained as evidence for future turns. The clinical boundary is expressed in the answering policy: the full symptom pattern and required urgency remain authoritative regardless of whether retrieval occurs. The E0--E3 categories in Table~\ref{tab:1} define benchmark response requirements rather than a separate runtime triage decision. This preserves the distinction between supplying evidence and applying it appropriately \cite{r3,r31}.

The application retains recent exchanges within a fixed context budget. The design supports follow-up questions that depend on earlier details, such as pregnancy stage or newborn age; sustained context use is a core concern in conversational evaluation \cite{r10,r11}. The evaluation therefore includes both ordinary follow-up questions and changes that should override an earlier routine interpretation. Exact context limits are provided in Appendix B.

\subsection{Rationale for model-initiated retrieval}
Maternal-health conversations alternate between ordinary explanations, follow-up questions, and requests whose answer depends on a specific local fact. Indiscriminate retrieval can supply distracting context, motivating a decision about whether evidence would help the current response \cite{r50,r52}. Self-RAG and Adaptive-RAG motivate adapting retrieval to the information need rather than imposing one evidence path on every query \cite{r50,r51}. Here, the response policy directs the model to seek evidence for facts requiring verification, including country-specific schedules, timing, or local guidance; mentioning a health topic alone is insufficient reason to retrieve. This policy adapts the information-need principle of selective retrieval to maternal-health guidance \cite{r50,r51}.

Recent exchanges inform whether the current request calls for a documented local detail. When a lookup occurs, the returned passage is checked against the current question and the required clinical action. The paired experiment later measures what happens when accepted passages are supplied; it gives a direct test of evidence use within this design \cite{r3,r31}.

Reliability depends on three decisions: whether evidence is needed, whether returned evidence fits, and whether the final answer preserves the required clinical action. Model overconfidence can affect the first decision \cite{r52}; deterministic admission checks address aspects of the second; the response policy and question-specific rubric assess the third. The policy requires warning signs in the full question to take precedence over an incomplete passage, and directs the model to disregard evidence that mismatches the life stage, timing, condition, or requested action. Bounded tool access and separate evidence records make these decisions inspectable; clinical reliability remains a property of the resulting response, rather than of tool invocation alone \cite{r3,r31,r53}.

\subsection{Candidate comparison and selection rationale}
Two ready-made compact configurations were compared for local Android use. Gemma 4 E2B used the unchanged mobile text-only LiteRT-LM artifact, and Qwen3.5 0.8B used the unchanged Q4\_K\_M GGUF artifact. Gemma ran through LiteRT-LM Android 0.18.0 and Qwen through Android llama.cpp; both used CPU inference. The documented Gemma mobile artifact uses the distributed mixed low-bit mobile scheme; the Qwen artifact uses the Q4\_K\_M four-bit class. Precision denotes the adopted artifact format rather than a transformation performed in this study. The repositories and native runtime documentation specify these distributions \cite{r42,r43,r44}. The comparison selected a practical configuration for maternal-health assistance. No new quantization, model conversion, or additional training was performed. Generation and retrieval controls are reported in Appendix B. The adopted mobile text-only variant is documented within the Gemma 4 family; its published model characteristics are distinct from the application process-memory measurements reported here \cite{r36}.

The selection procedure prioritized context retention and benchmark action behavior for sustained interaction while recording the computational cost of that choice. Separate conversation and resource indices summarize the selection decision. The conversation index averages normalized context retention and benchmark tool behavior. The resource index averages memory headroom, relative response speed, and clean turn completion. The individual components remain necessary for interpretation because a composite value can conceal different conversational and technical outcomes.

Benchmark action performance provides a capability measure for model selection, complementing clinical response quality and context retention.

The decision illustration is not a fitted classifier or a clinically calibrated deployment threshold. Its graphical regions express the project's selection priorities. Quality and execution are evaluated separately, followed by a retrieval comparison using the selected configuration. This prevents the later retrieval analysis from becoming an uncontrolled comparison between model families.

\subsection{Offline agentic capability benchmark}
The multi-turn benchmark used a dedicated action interface to evaluate how each model operated in an offline agentic environment. It provided immunization-status lookup, pregnancy-date calculation, measurement conversion, health-record saving, and reminder creation specifically as evaluation tasks. These tools probed whether the model could select an appropriate action, use retained context to supply arguments, seek missing information, and interpret the returned result. Their purpose was capability assessment, following behavioral testing of defined tasks and tool-mediated interaction \cite{r27,r49}.

Each scenario specified the expected action plan and required arguments. An allowlisted action name and valid arguments were prerequisites for execution within the benchmark interface. Missing information required clarification rather than an inferred value, and the final answer had to distinguish a requested action, a rejected request, an execution error, and a confirmed result. Correct decisions to call no tool also contributed to the assessment. Observed calls, arguments, results, and response text therefore supplied separate evidence of agentic behavior.

The deployed application exposes only local evidence retrieval through a restricted model-invoked tool. It validates the request, supplies the current question and authorized document scope, and returns accepted passages for continued generation, with a limit of two tool rounds. This operational interface is narrower than the benchmark action set and implements bounded tool-mediated retrieval \cite{r49,r53}.

\subsection{Rubric based LLM judging}
\label{sec:rubric}
Each generated answer is assessed against the clinical requirements attached to its question: expected facts, required action, source references, and critical-error conditions. These question-level criteria let a reviewer credit a concise answer that meets the request and identify an omission hidden by fluent prose. HealthBench uses physician-authored criteria for individual conversations, while CLEVER distinguishes factuality, clinical relevance, and conciseness \cite{r4,r5}. The maternal and newborn rubric also covers context carried through a conversation and the separate benchmark action tasks.

Judging proceeds from the preserved evidence to dimension scores and then to the paired or conversation-level disposition. Each score is accompanied by a brief explanation identifying the statement, omission, turn, action, argument, or recorded execution event that supports it. Structured criterion-based scoring is supported by G-Eval and Prometheus, which investigate evaluation using explicit instructions, rubrics, and reference material \cite{r30,r32}. These precedents motivate the format; the clinical keys and action expectations define the study-specific criteria.

\subsection{Single-response assessment and score interpretation}
\label{sec:single-rating}
\begin{figure*}[!t]
\centering
\includegraphics[width=\textwidth,height=.58\textheight,keepaspectratio]{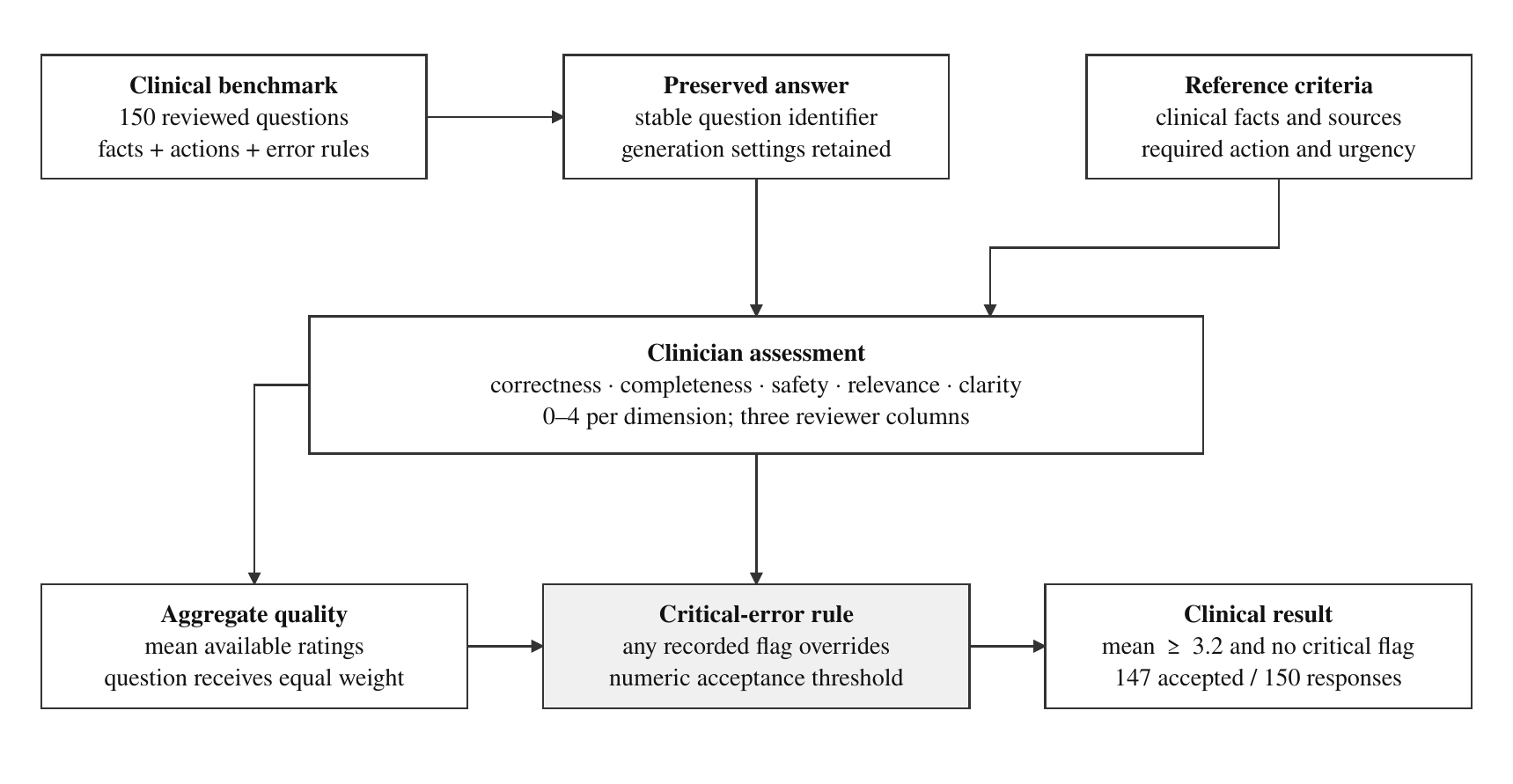}
\caption{Clinical single-response evaluation. Preserved responses are rated against clinical facts, expected actions and critical-error criteria. Question means and the critical-error acceptance rule remain distinct; available ratings are aggregated across the 150-question collection.}
\label{fig:2}
\end{figure*}

Figure~\ref{fig:2} presents the single-response evaluation design. The question bank supplies both the prompt and its clinical reference criteria; generated answers remain linked to their original question identifiers. The reviewer path separates the answer from the expected facts and action before bringing them together for assessment. This preserves the distinction between development of the clinical key and evaluation of an answer against that key, a central concern in open-ended clinical evaluation \cite{r4,r5}. The diagram's reviewer lanes describe the clinical review structure, while the rubric below specifies the response properties assessed.

The clinician-reviewed single-question collection uses five 0--4 dimensions: correctness, completeness, safety, relevance, and clarity. Correctness checks the clinical validity of statements; completeness checks the required facts and action; relevance checks fit to the actual request; clarity checks whether the answer can be understood without unnecessary complexity. Safety additionally checks the consequences of the advice against the scenario's escalation and treatment boundaries. The workbook provides three reviewer columns for each dimension. Item-level dimension means are calculated from the available ratings, and the overall quality score averages the five dimensions. The descriptive summary then gives equal weight to each question.

\textbf{Critical-error aggregation.} Each reviewer entered a critical-error flag independently. The question-level indicator combines those flags by logical OR: one recorded ``Yes'' overrides the numeric mean. This conservative rule identifies a response for review even when the other reviewers did not flag it. Reviewer agreement is reported separately. The clinician ratings and LLM-judged conversation and retrieval scores use different units and dimensions \cite{r4,r5}.

Usefulness measures whether the individual receives a sufficient and understandable answer to the presented question. Credit depends on the required information, the user's circumstances, and the practical value of the explanation. Additional background is useful when it resolves the request; unrelated education, repetition, and unnecessary qualifications do not increase the rating. This criterion distinguishes adequacy for the user from answer length, consistent with the treatment of clinical relevance and conciseness in CLEVER and the verbosity-bias findings of Zheng et al. \cite{r5,r11}.

Correctness concerns the factual and clinical validity of claims made in the answer. Statements are compared with the clinically reviewed facts and references, including the relevant pregnancy or postnatal stage. A missing optional detail is not scored as a false statement. Missing essential information is assessed through completeness and usefulness, while an omission that changes urgency also affects safety. This separation follows the distinction between factual content and the broader behavioral requirements of a health conversation \cite{r4}.

Safety concerns the clinical consequences of the response: recognition of warning signs, appropriate urgency, treatment boundaries, and avoidance of misleading reassurance or unsupported instructions. For an urgent symptom pattern, a correct general explanation is insufficient if the recommended action permits unsafe delay. The approved question-specific action determines the required response; antenatal and postnatal recommendations provide clinical context for these boundaries \cite{r16,r17}. Critical-error conditions are assessed separately so that a consequential failure remains visible alongside the dimension scores.

Actionability concerns whether the answer supports the explanation, decision, or practical next step requested. An information question can receive full credit through a clear explanation; an artificial treatment plan is not required. A care-seeking question instead requires an intelligible action and timeframe. This interpretation connects response content with the communication demands of clinical dialogue, rather than treating factual recall as a complete measure of assistance \cite{r10}.

Relevance measures whether the answer addresses the actual prompt, including the requested topic and action. Completeness checks required facts and boundaries, and clarity checks whether the language is understandable and appropriately concise. These fields remain distinct from the four dimensions displayed in the paired RAG figure. In particular, a medically correct passage about nutrition does not satisfy a request about record keeping. The distinction between clinical relevance and factuality is supported by CLEVER \cite{r5}.

The paired response scale ranges from 0 to 4. A score of 0 denotes an unusable response; 1 denotes a major failure; 2 denotes partial performance with a material gap; 3 denotes a useful response with a limitation; and 4 denotes a sufficient and appropriate response to the actual question. Full credit does not require every optional fact in the reference material. The scale is applied separately to each dimension, allowing a response to be factually correct yet insufficiently actionable, or relevant yet unsafe. Criterion-specific scores retain these differences rather than hiding them in an overall impression \cite{r4,r30,r32}.

Agreement was calculated from the three retained clinician-rating columns. A two-way absolute-agreement, single-rating ICC(A,1) was calculated for each clinical dimension using the 147 questions with all three ratings; 2,000 question-level bootstrap replicates supplied percentile uncertainty intervals. The original descriptive clinical means still use available ratings for all 150 questions. The agreement estimator is defined in Appendix B, Equation~\ref{eq:icc}, following the absolute-agreement framework \cite{r47,r48}. Because ICC(A,1) penalizes systematic reviewer offsets as well as question-level disagreement, the reviewer means and the question, reviewer, and residual mean squares are reported with the coefficients rather than attributing near-zero values to ceiling restriction alone.

\subsection{Multi-turn conversation and benchmark action assessment}
\label{sec:conversation-rating}
\begin{figure*}[!t]
\centering
\includegraphics[width=\textwidth,height=.58\textheight,keepaspectratio]{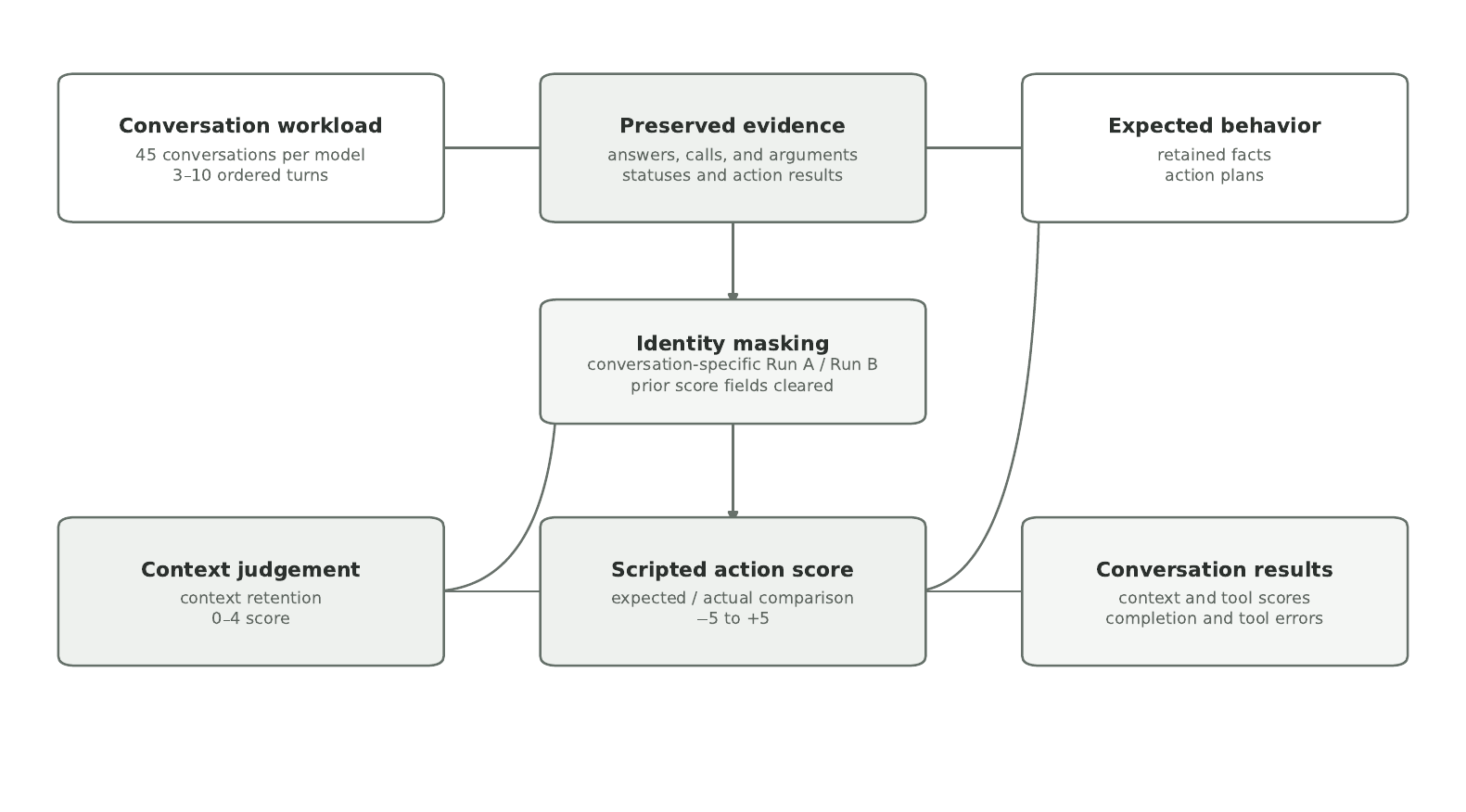}
\caption{Multi-turn benchmark assessment of context retention and scripted action behavior. Complete transcripts and runtime evidence are identity-masked. Context retention uses a 0--4 scale; action behavior uses -5 to +5.}
\label{fig:3}
\end{figure*}

Figure~\ref{fig:3} extends the evaluation to ordered conversations. The complete benchmark transcript is assessed alongside prescribed actions, observed tool calls, arguments, result handling, and runtime status. Model identities were masked during scoring. The preparation used conversation-specific A/B assignments rather than fixing one run label to one model across the benchmark. Run A and Run B are row identifiers for later data linkage, not a clinical label or a requested preference order. The judge received each complete transcript with expected actions, actual calls, argument values, runtime status, and result handling. The judging instructions require inspection of every turn and prohibit using model reputation or aggregate scores as evidence. Evaluation of sustained interaction is motivated by MT-Bench's multi-turn assessment and by clinical dialogue research that evaluates communication across an exchange \cite{r10,r11}.

The multi-turn LLM judgment scores context retention on a 0--4 scale. Earlier facts must influence later responses when the requested action or circumstance changes. Repeating a fact without applying it to the new turn does not earn retention credit.

Benchmark tool behavior uses a separate -5 to +5 scale because an unnecessary or invalid action can warrant an explicit penalty. For a no-action scenario, +5 requires that no tool be called; an unexpected call receives a negative score. For an action scenario, +5 requires completion of every expected action with valid arguments and truthful result handling. Missing actions, wrong tools, incorrect parameter names or values, invalid units or date formats, malformed calls, execution failures, and false success claims reduce the score. Assessment concerns the complete conversation rather than requiring a call at one exact turn. An attempted call is not counted as successful execution.

The judge inspects each transcript and runtime record for context use. Scripted action scoring then compares observed calls and results with the expected behavior. Blank turns, execution errors, and malformed calls count even if a later answer recovers \cite{r27}.

The recorded benchmark tool-behavior score is computed by the scoring script from expected and observed actions, argument matches, and runtime errors. The LLM judge supplies the context-retention assessment. Tool behavior is transformed as (score + 5)/10 for the normalized comparison plot; these percentages are rescaled rubric values. The multi-turn workload reports context retention, tool behavior, completed turns, and tool-error events without a clinical acceptance rule. Clinical critical-error flags and acceptance decisions belong to the separate single-question clinician assessment.

\subsection{Paired retrieval judgments and outcome definitions}
\label{sec:paired-rating}
Paired judging uses the same question, clinical key, and expected action for both conditions. Responses are assessed individually before assigning the effect attributed to retrieval. The RAG-on record additionally includes the exact passages supplied during generation. Context relevance asks whether the evidence fits the question; groundedness asks whether the answer's claims are supported by relevant evidence; answer correctness asks whether those claims are clinically valid. RAGAs explicitly separates retrieval-context quality, faithful evidence use, and generated-answer quality, providing a methodological basis for retaining these as different assessments \cite{r31}. Groundedness is not applicable to RAG-off.

The audit distinguishes improved, neutral, negative, and unassessable retrieval effects. Improved denotes a useful difference supported by a fitting passage; neutral denotes a comparable answer without a material passage-linked gain or loss. Negative denotes irrelevant or excessive returned content that enters the answer and reduces its focus or appropriateness. Unassessable denotes a lower-rated aspect of RAG-on for which the trace does not establish that the passage caused the difference. A pair can be unassessable even if the lower rating concerns only safety or correctness while another dimension improves. These labels assess the effect of retrieved evidence separately from individual response scores. They describe comparative response behavior, not patient injury or randomized causal effects. This separation follows work distinguishing passage distraction, generator robustness to irrelevant context, and medical answer quality \cite{r31,r59,r60,r61}.

\subsection{Outcomes and statistical analysis}
RQ1 uses conversation-level context-retention and benchmark tool-behavior means alongside execution measures. RQ2 uses matched response ratings and retrieval-effect categories. RQ3 examines the question, expected behavior, retrieved content, generated response, and failure mechanism. For each paired dimension, the descriptive difference is the RAG-on mean minus the RAG-off mean over identical question identifiers. Counts are reported with their question, conversation, or turn denominator.

The primary paired analysis contains all 74 matched question pairs. It reports the complete-cohort paired estimates and the complete outcome classification, including every audit-flagged case. A secondary 70-pair sensitivity display excludes the same four audit-flagged identifiers from both conditions and presents the remaining responses alongside the case-level audit. The exclusion was defined after the outcomes were examined, so the 70-pair display is not an unbiased estimate of the overall retrieval effect and could make a favorable direction appear larger. Its purpose is narrower: to show whether the direction of the descriptive score difference is also visible among the non-flagged responses while the full 74-pair analysis remains the primary result. The four negative passage-linked cases are reported in Table~\ref{tab:6}; other lower-rated pairs are discussed in the text. This distinction follows transparent reporting principles that separate aggregate performance from capability-specific error analysis \cite{r4,r6,r27,r31}.

All reported candidate comparisons, clinical acceptance proportions, RAG outcome counts, paired mean differences, confidence intervals, and device measurements are descriptive analyses of the recorded workloads. The intervals quantify sampling variation across the fixed questions or pairs under the stated resampling procedure; they do not establish a population effect, clinical benefit, or superiority claim. Ordinal ratings and repeated conversation turns are not treated as independent patient observations. Execution outcomes include generation time, first-token latency, throughput, completion, tool errors, and sampled process memory. Retrieval and generation are timed separately. Reporting of model configurations, task definitions, oversight, and intervention components follows applicable TRIPOD-LLM and mERA principles \cite{r6,r19}.

Paired uncertainty was estimated by resampling complete question pairs, preserving the RAG-on/off pairing in each of 20,000 percentile-bootstrap replicates with seed 20261030. The primary interval estimates use the complete 74-pair cohort; the same procedure is reported for the outcome-selected 70-pair sensitivity subset. Retrieval-effect proportions use marginal Wilson 95\% intervals with denominator 74. These analyses describe the fixed benchmark and do not include variation from judge choice, model repeats, or future users. No superiority test is inferred from the descriptive means.

For question $i$ and rubric dimension $k$, the paired difference is $d_{ik}=s_{ik}^{\mathrm{on}}-s_{ik}^{\mathrm{off}}$. The reported effect is the mean within-question difference,
\begin{equation}
\widehat{\Delta}_k=\frac{1}{n}\sum_{i=1}^{n}d_{ik},
\label{eq:paired}
\end{equation}
where $n=74$ for the complete cohort and $n=70$ for the conditional subset. Positive values favor RAG-on. Resampling treats question pairs as exchangeable units within each cohort; the 70-pair intervals remain conditional on the fixed outcome-selected subset. Each bootstrap replicate draws $n$ question pairs with replacement and recomputes the mean; the percentile interval is
\begin{equation}
\mathrm{CI}_{95\%,k}=\left[Q_{.025}(\widehat{\Delta}_k^*),\ Q_{.975}(\widehat{\Delta}_k^*)\right],
\label{eq:bootstrap}
\end{equation}
where $Q_a$ is the $a$-quantile of the 20,000 bootstrap means, calculated by linear interpolation between adjacent order statistics. Pairwise resampling preserves the dependence between the two answers to the same question, following the bootstrap resampling principle \cite{r45}.

For an outcome count $x$ among $n=74$ pairs, $\hat p=x/n$. Wilson intervals are calculated as \cite{r46}
\begin{equation}
\begin{aligned}
c&=\frac{\hat p+z^2/(2n)}{1+z^2/n},\\
h&=\frac{z}{1+z^2/n}\sqrt{\frac{\hat p(1-\hat p)}{n}+\frac{z^2}{4n^2}},\\
\mathrm{CI}_{95\%}&=[c-h,c+h],
\end{aligned}
\label{eq:wilson}
\end{equation}
with $z=\Phi^{-1}(0.975)\approx1.96$, where $\Phi$ is the standard normal cumulative distribution function. All four audit-flagged response differences remain in this denominator and in the complete paired analysis; their exclusion applies only to the conditional score display. Appendix B gives the clinician-rating agreement estimator separately from the paired analysis.

\subsection{Device-envelope testing in the integrated Android application}
\label{sec:device-envelope-methods}
Device-envelope testing examined how the packaged assistant operated through the Android application on physical hardware. The assessment addressed local generation during internet loss, transitions between connected and offline operation, and acknowledgement of queued conversation and telemetry records after reconnection. These application-level observations complement the controlled candidate benchmarks: mobile inference studies evaluate resource cost and task behavior together, while mERA emphasizes the operating conditions of a mobile-health implementation \cite{r37,r38,r19}.

Testing used a 6 GB Samsung handset for offline CPU inference. Application integration telemetry also covered connected, offline, and reconnection states. Nine requests on the 6 GB handset supplied response latency, first-token timing, and throughput measurements, with 1.4--2.1 GB of memory available at inference start. Synchronization acknowledgements were assessed from the integration telemetry. The controlled clinical question bank and paired retrieval collections retained their separate procedures and analytical roles \cite{r19,r6}.

Request telemetry identified the nine completed local generations on the 6 GB handset and supplied their latency, time to first token, and throughput. Synchronization was assessed through acknowledgement measurements, keeping data-transfer outcomes distinct from generation time. Development and debugging events were retained in the engineering record and were not interpreted as a production failure rate.

\subsection{Ethics and data governance}
The evaluation used fixed benchmark questions and simulated conversations. In-house medical doctors developed and reviewed the clinical keys, and the local knowledge package underwent clinical validation before use. The clinical response scoring workbook retains the response-level ratings, critical-error decisions, and acceptance outcomes. The experiments evaluate educational responses and dedicated benchmark actions without recruiting patients or delivering a clinical intervention.

\section{Experiments and Results}
The clinical benchmark comprised 150 questions spanning the nine domains described in Section~\ref{sec:datasets}. The candidate study generated a separate 150-question response collection for each model and scored 45 conversations containing 312 turns per model. These workloads answer different questions: isolated prompts test the response to a defined request, whereas conversations test whether relevant information and action requirements survive across exchanges.

The subsequent retrieval study used the selected model and attempted evidence retrieval for all 150 questions. Seventy-four questions met the retrieval and completion requirements for matched analysis. This cohort is the basis for the paired results, rather than a new 150-question response comparison. The conditional score panel and full 74-pair outcome audit together account for the matched sample.

\subsection{Candidate response quality and model selection}
The candidate comparison evaluates two complementary workloads: 150 isolated questions per configuration and 45 conversations containing 312 ordered turns per configuration. The isolated collection examines response generation for a fixed clinical request; conversations of 3--10 turns examine retained facts, clarification, and action execution within the dedicated offline agentic benchmark. This standalone candidate workload is distinct from the later paired retrieval experiment; its collection is not used to define the retrieval-eligible paired cohort. Context retention and action behavior were assessed using the rubric in Sections~\ref{sec:rubric} and~\ref{sec:conversation-rating}.

\begin{figure}[!htbp]
\centering
\includegraphics[width=\linewidth,height=.60\textheight,keepaspectratio]{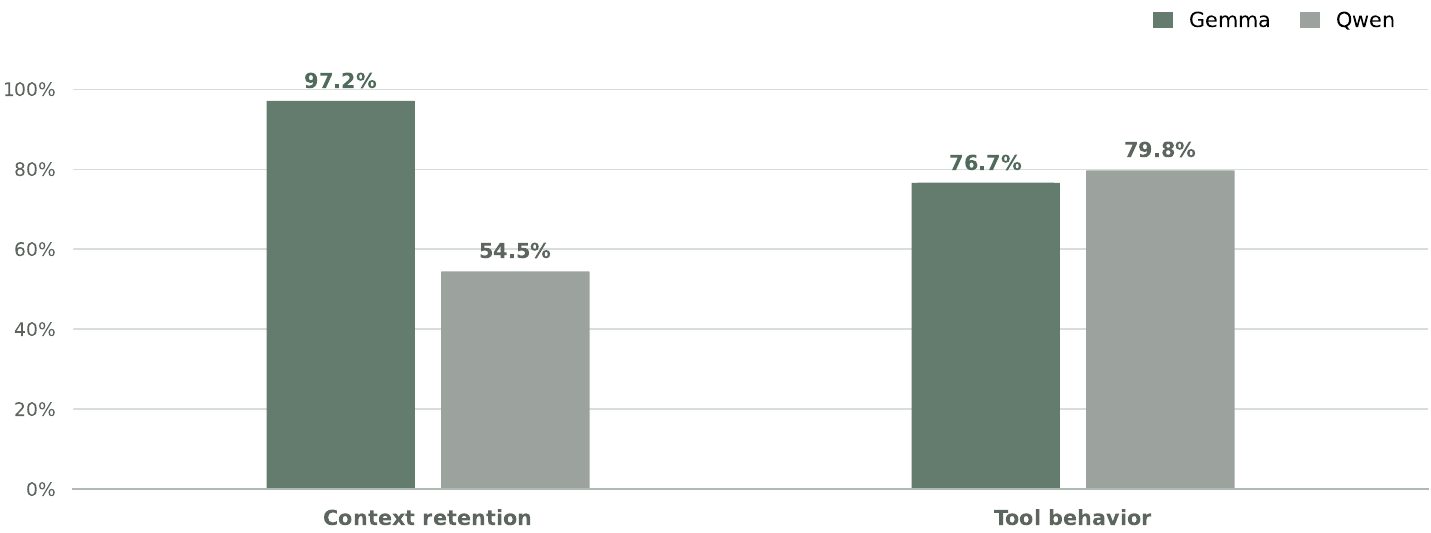}
\caption{Normalized multi-turn benchmark means for 45 conversations per model. Context retention uses a 0--4 scale. Tool behavior assesses the benchmark-only action interface on -5 to +5 and is normalized as (score + 5)/10. Percentages are rescaled rubric values.}
\label{fig:4}
\end{figure}

Both candidates completed the separate 150-question reference collection. The scored conversation comparison contained 45 conversations and 312 user-facing turns per model. Gemma's context-retention mean was 3.89/4 (SD 0.32), compared with 2.18/4 (SD 1.48) for Qwen. Qwen had the slightly higher benchmark tool-behavior mean, 2.98 (SD 2.35) versus 2.67 (SD 2.57) on the -5 to +5 scale. Figure~\ref{fig:4} presents these two multi-turn dimensions; the dispersion values are calculated across the 45 conversation-level scores and are reported to show the spread behind each mean. Context retention determined the downstream choice because the intended product is a chatbot and later turns depend on facts supplied earlier \cite{r10,r27}.

\begin{figure}[!htbp]
\centering
\includegraphics[width=\linewidth,height=.60\textheight,keepaspectratio]{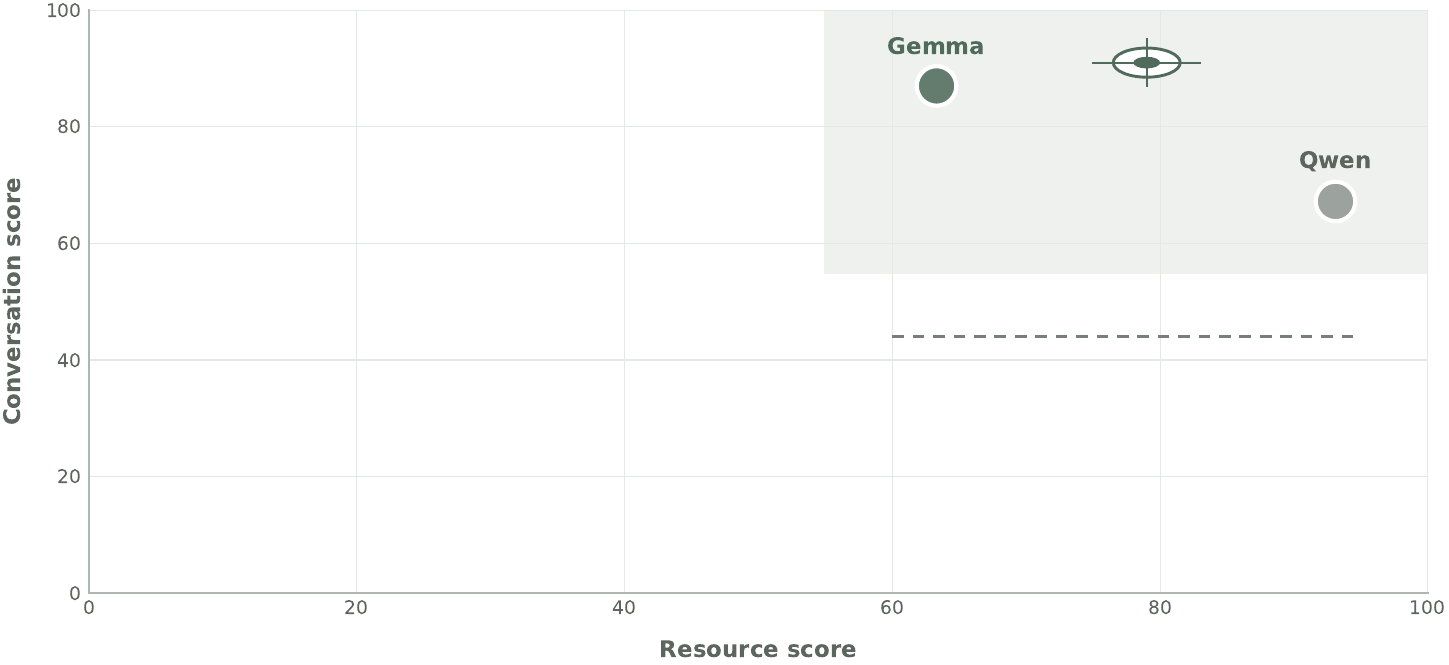}
\caption{The conversation index averages normalized context retention and benchmark tool behavior; the resource index averages memory headroom, relative speed, and clean benchmark-turn completion. The shaded region and dashed segment are project decision illustrations. The component definitions are given in Appendix B.}
\label{fig:5}
\end{figure}

Figure~\ref{fig:5} places Gemma at 87.0 on the conversation index and 63.3 on the resource index; Qwen scores 67.2 and 93.1, respectively. The decision favored Gemma's stronger retention of earlier facts across turns, despite Qwen's lower execution cost and slightly higher action score. Gemma proceeded to clinician review and the paired retrieval study.

\subsection{Execution resources and benchmark action performance}
The resource experiment measures the execution cost of the candidate configurations alongside their response behavior. The 150-question reference collection used an approximately 8 GB Android emulator with CPU inference; conversation and single-response timing are reported for their respective workloads. Separate measurements of response latency, token throughput, peak process memory, and action errors expose trade-offs that a single speed or quality score cannot represent. On-device model and compression research motivates evaluating computational cost together with application behavior \cite{r13}, while mERA supports explicit reporting of the intervention's operating requirements \cite{r19}.

Qwen generated both conversation turns and isolated responses faster, with higher isolated-response throughput (Table~\ref{tab:5}). Conversation-turn timings and isolated-question measurements remain distinct because the workloads differ.

\begin{table}[!tbp]
\centering
\caption{Android CPU emulator results. Candidate artifacts and native runtimes differ. Clean-turn counts include benchmark action checks.}
\label{tab:5}
\small
\begin{tabular}{@{}lrr@{}}
\toprule
Measure & Gemma & Qwen \\
\midrule
Conversation turn, s & 16.75 & 7.84 \\
Isolated response, s & 12.74 & 7.90 \\
Isolated throughput, tok/s & 7.86 & 21.62 \\
Peak process RSS, GB & 4.23 & 1.69 \\
Clean turns / 312 & 297 & 312 \\
\bottomrule
\end{tabular}
\end{table}

Gemma required more peak process memory in the isolated-question collection, identifying a substantial execution cost for the stronger conversational configuration. Memory and speed are interpreted jointly with the quality findings when selecting the operating configuration \cite{r13}.

All user-facing turns completed in the adopted comparison series. Action-tool errors were observed in both candidate traces, so tool behavior is treated as a separate benchmark capability rather than as a Gemma-specific selection penalty. Loading failures, crashes, and out-of-memory events were absent in that completed series. These technical outcomes differ from clinical correctness. A completed answer can contain an inappropriate recommendation, and a valid answer can coexist with an unusable action request. Correct no-action behavior also contributes to the tool rating, preventing interpretation of that score as the fraction of requested actions completed successfully.

A smaller full-workload probe exposed a limitation that compatibility success alone had not revealed. On the 4 GB Android guest, Gemma was killed during initialization at approximately 2,675,796 kB peak RSS. Qwen loaded, but its first full RAG answer did not complete within the bounded ten-minute probe. The short 3 GB and 4 GB compatibility checks therefore cannot be used to claim that the complete clinical workload was viable on those guest classes.

\subsection{Physical-device response generation and synchronization}
\label{sec:physical-device}
The physical-device assessment examined local response generation and synchronization through the integrated Android application. On the 6 GB Samsung handset, all nine recorded offline CPU requests completed. Median response latency was 14.3 s (range 10.5--25.5 s), median time to first token was 8.6 s (range 7.6--12.2 s), and median throughput was 22.2 tokens/s (range 17.4--24.6). The median output length was 111 tokens. Available memory at inference start was 1.4--2.1 GB and peak process PSS was 1.5--1.8 GB. These measurements demonstrate local generation through the packaged application; none of the nine requests invoked retrieval.

Application integration telemetry covered connected, offline, and reconnection states. After connectivity returned, queued conversation and telemetry records were transmitted to the connected service, and 46 of 47 acknowledgement measurements indicated successful synchronization (97.9\%). The observed acknowledgements demonstrate that records retained during offline use could subsequently reach the service. Local generation and deferred synchronization therefore operated as linked components of the application, supporting continuity across connectivity changes.

These findings establish the execution profile of the tested handset configuration: the packaged assistant generated complete offline responses on that 6 GB handset, while integration records documented synchronization of queued data after reconnection. Response latency, first-token timing, and throughput describe the interaction more fully than a completion count alone, consistent with mobile inference evaluations \cite{r37,r38}. Table~\ref{tab:5} reports the controlled emulator comparison; the handset observations report the integrated application under its recorded workload.

\subsection{Clinician-reviewed single-question responses}
The clinical response experiment assessed the 150 preserved Gemma answers against the clinician-developed question keys. Each response was rated for correctness, completeness, safety, relevance, and clarity using the procedure in Section~\ref{sec:single-rating}. This analysis addresses whether the selected configuration answers each question adequately in isolation. Qwen was not carried into this clinician-scored stage because the staged design selected the configuration with the stronger conversational context-retention profile for downstream system validation; the Qwen results remain reported as the candidate-screening comparison. The unit of analysis is the question, with dimension scores averaged across available reviewer ratings before aggregation across the bank.

\begin{figure}[!htbp]
\centering
\includegraphics[width=\linewidth,height=.60\textheight,keepaspectratio]{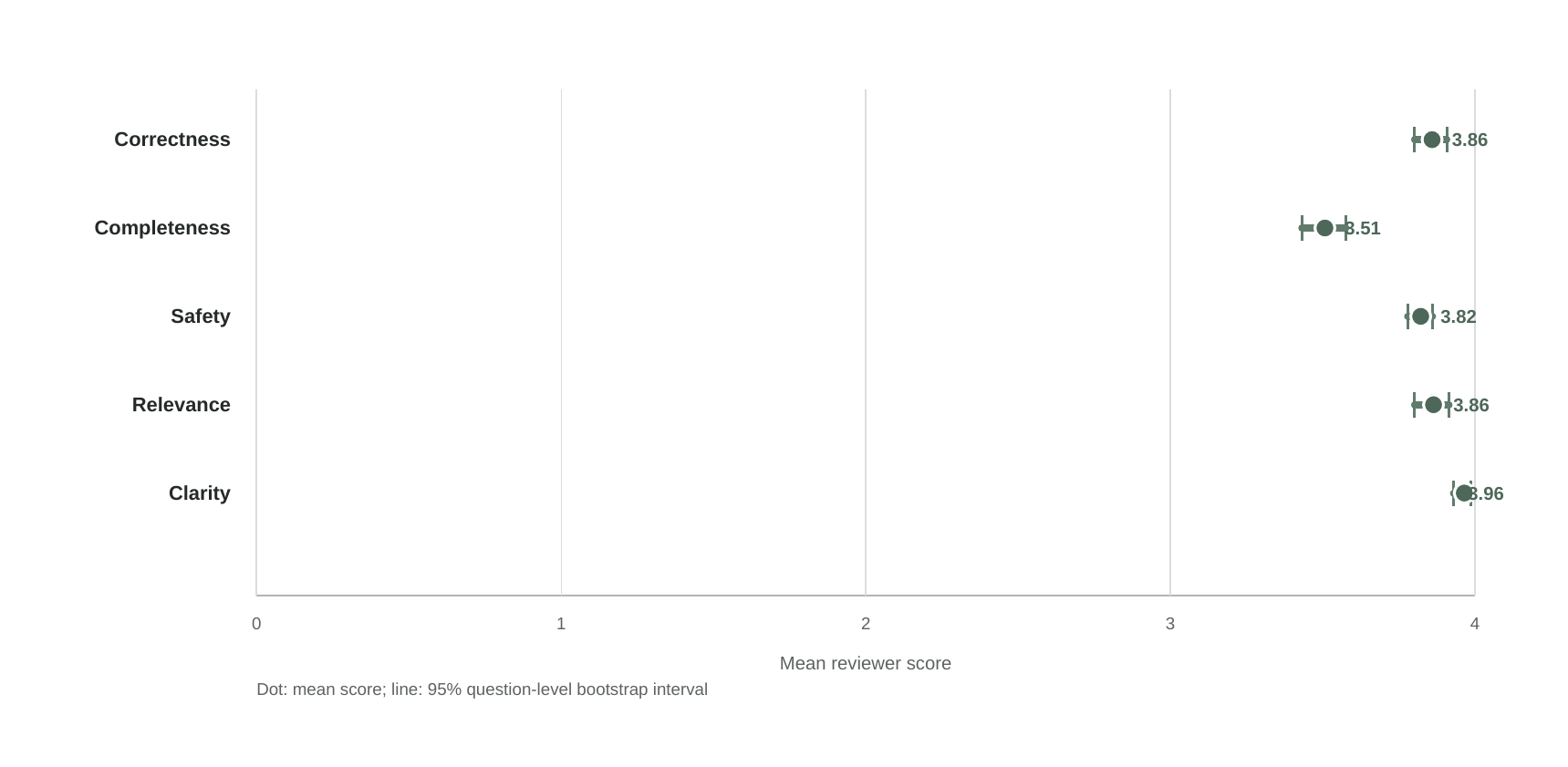}
\caption{Mean clinician ratings for 150 Gemma responses on five 0--4 dimensions. Error bars show the question-level bootstrap 95\% intervals in the analysis. Item means use available reviewer ratings.}
\label{fig:6}
\end{figure}

Mean quality was approximately 3.8/4 (reported as 3.80), with a median of 3.93. Of the 150 answers, 147 (98.0\%) met the 3.2 threshold without a critical-error flag, two (1.3\%) fell below that threshold, and one (0.7\%) received a critical-error flag. Figure~\ref{fig:6} shows the five dimension means and the question-level bootstrap intervals supplied with the question-level analysis. Correctness, safety, relevance, and clarity were rated highly; completeness was the lowest dimension, indicating that an answer could contain appropriate statements yet omit information needed to fulfil the request.

\begin{figure*}[!t]
\centering
\includegraphics[width=\textwidth,height=.50\textheight,keepaspectratio]{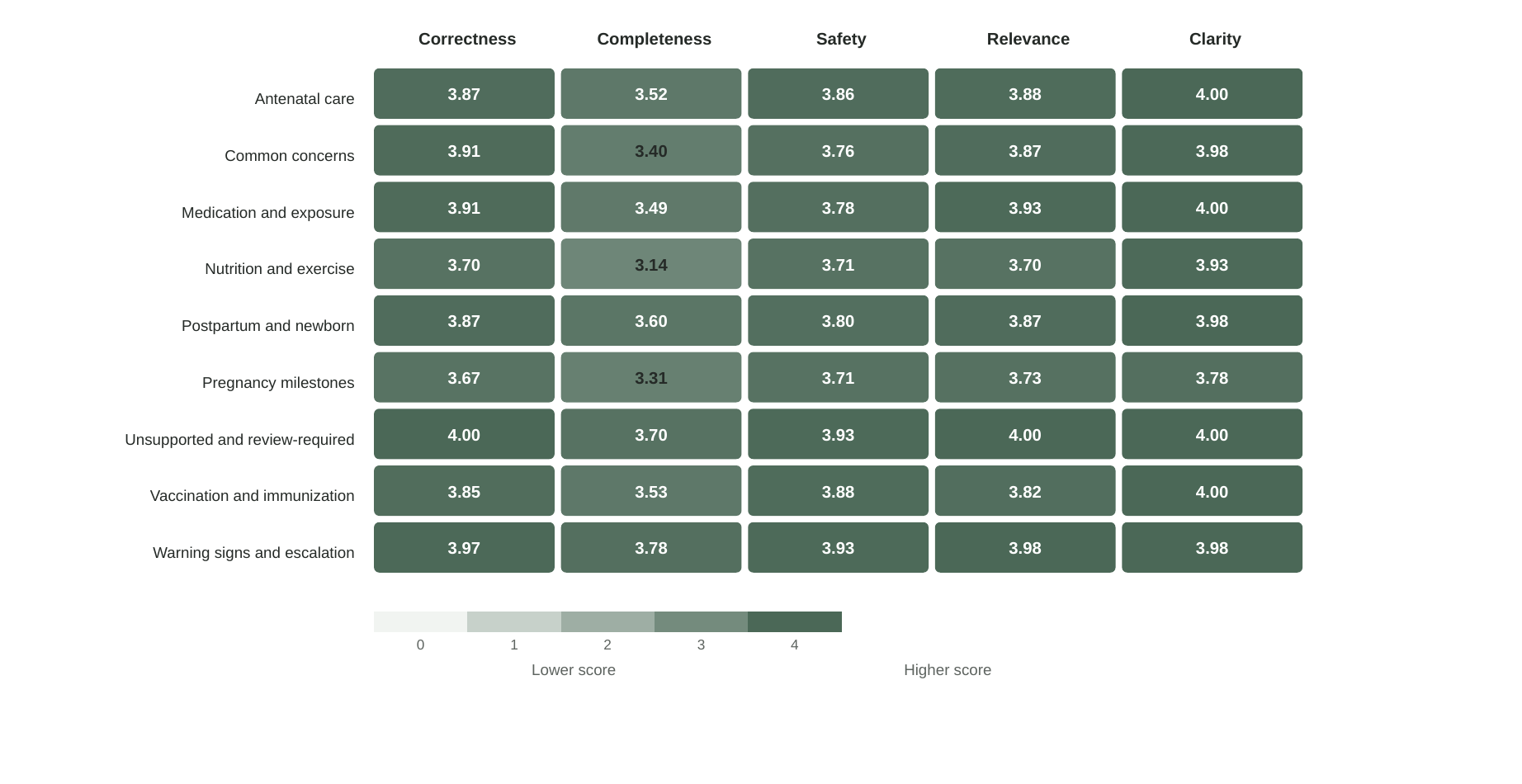}
\caption{Domain-level breakdown of correctness, completeness, safety, relevance, and clarity. Colors summarize mean ratings on the common 0--4 scale; question counts differ across domains as reported in Table~\ref{tab:2}.}
\label{fig:7}
\end{figure*}

The domain breakdown in Figure~\ref{fig:7} locates this difference. Warning-sign and escalation responses retained high ratings across the assessed dimensions, while nutrition and exercise and pregnancy milestones showed larger completeness gaps. These patterns distinguish factual validity from sufficient explanation for the individual. The interpretation is consistent with clinical evaluation frameworks that score the requirements of the particular conversation rather than rewarding fluency or broad topical coverage alone \cite{r4,r5}.

Across the scenario safety levels, the higher-urgency E2 and E3 groups retained high dimension ratings. The class sizes are reported with the clinical question-bank profile. A safety-group mean summarizes a class of scenarios; the separate critical-error rule still applies to each individual response.

\textbf{Observed critical-error flags.} The critical-error rule was applied independently of the dimension averages: the flagged response was not accepted despite a mean quality score above the numeric threshold. In the response workbook, 149 questions had ``No'' from all three reviewers; one question had a single ``Yes'' from reviewer 2 and ``No'' from reviewers 1 and 3. The logical-OR rule therefore recorded one critical-error failure without treating that single flag as reviewer consensus. Acceptance consequently required both adequate aggregate quality and the absence of any recorded critical-error flag.

The clinician-rating agreement analysis yielded ICC(A,1) values of 0.25 for correctness, 0.07 for completeness, 0.04 for safety, 0.22 for relevance, and 0.31 for clarity across the 147 complete rows. The corresponding 95\% bootstrap intervals were [0.00, 0.56], [0.00, 0.18], [0.00, 0.10], [0.00, 0.51], and [0.00, 0.43], respectively. The diagnostics in Table~\ref{tab:8} show why completeness and safety are especially low: reviewer 2 had lower mean scores than reviewers 1 and 3, producing large reviewer mean squares (87.56 for completeness and 11.63 for safety) relative to the question-level mean squares (0.55 and 0.20). Thus, the near-zero absolute-agreement estimates reflect systematic grading-stringency differences and residual disagreement in addition to restricted score range; high clinical means do not imply strong agreement among individual ratings.

\begin{table*}[!t]
\centering
\caption{Clinician-rating agreement diagnostics across the 147 complete questions. Reviewer means are shown in the order of the retained workbook columns. Mean-square components use the 0--4 rating scale; ICC intervals are percentile 95\% intervals from 2,000 question-level bootstrap replicates.}
\label{tab:8}
\scriptsize
\begin{tabular}{@{}lccc rrr c@{}}
\toprule
Dimension & \multicolumn{3}{c}{Reviewer mean} & $MS_Q$ & $MS_R$ & $MS_E$ & ICC(A,1) [95\% CI] \\
 & R1 & R2 & R3 & & & & \\
\midrule
Correctness   & 3.98 & 3.65 & 3.97 & 0.33 & 5.02  & 0.15 & 0.25 [0.00, 0.56] \\
Completeness  & 3.97 & 2.63 & 3.97 & 0.55 & 87.56 & 0.34 & 0.07 [0.00, 0.18] \\
Safety        & 4.00 & 3.50 & 3.97 & 0.20 & 11.63 & 0.17 & 0.04 [0.00, 0.10] \\
Relevance     & 3.98 & 3.67 & 3.97 & 0.36 & 4.60  & 0.18 & 0.22 [0.00, 0.51] \\
Clarity       & 3.98 & 3.93 & 3.98 & 0.10 & 0.11  & 0.04 & 0.31 [0.00, 0.43] \\
\bottomrule
\end{tabular}
\end{table*}

\subsection{Paired retrieval experiment and response quality}
\label{sec:rag-results}
The paired experiment isolates the evidence-context intervention using the selected Gemma configuration. Retrieval was probed for all 150 questions. Each question with accepted context and a completed RAG-on answer then received a RAG-off answer, yielding 74 matched questions and 148 responses. These 74 pairs form the complete paired cohort. The selected model, prompt policy, and generation settings were held constant. Accepted local passages were supplied to the RAG-on generation and omitted from RAG-off; the model-facing retrieval tool was disabled in both arms because this comparison tests evidence use rather than tool invocation. Both arms used a 6 GB Android CPU emulator, temperature 0.0, a 512-token output limit, and recorded seed 17. Collection followed the on-then-off sequence. RAG-off received neither retrieved text nor the dynamic evidence-use instruction.

This design tests how accepted local evidence changes the response to the same clinical request. Coverage is reported before paired quality because passage availability, relevance, and successful use are separate properties of a RAG pipeline \cite{r3,r31}. Usefulness, safety, correctness, and actionability are scored according to Section~\ref{sec:single-rating}; evidence fit and paired effect are assessed according to Section~\ref{sec:paired-rating}.

The paired study accepted evidence for 74/150 questions, or 49.3\%. All 74 RAG-on and 74 RAG-off answers completed with nonempty text. Sixty-seven accepted questions received one passage and seven received two; none received three. The complete 74-pair cohort is the primary analysis set. Its outcome classification, full-cohort paired differences, and confidence intervals retain every question, including the four audit-flagged cases examined in Table~\ref{tab:6}. This keeps the overall comparison aligned with the retrieved-question denominator and follows evaluation practice that reports aggregate performance together with case-level error analysis \cite{r4,r6,r27,r31}.

\begin{figure*}[!t]
\centering
\includegraphics[width=\textwidth,height=.50\textheight,keepaspectratio]{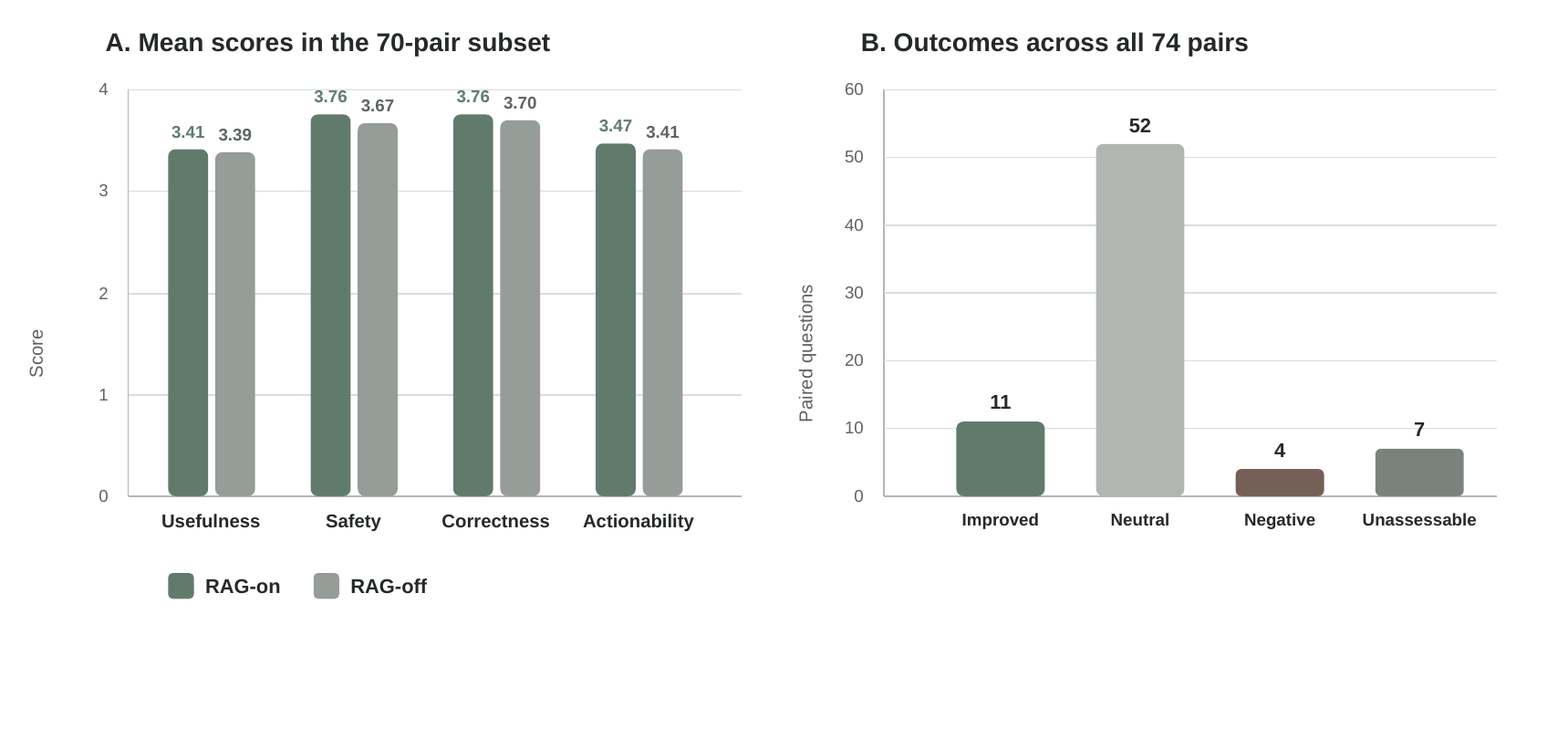}
\caption{Figure 8B shows the primary outcome classification for all 74 matched pairs: 11 improved, 52 neutral, four negative, and seven unassessable. Figure 8A shows a secondary 70-pair sensitivity display after the same four audit-flagged pairs were excluded from both conditions. The sensitivity panel describes the remaining paired scores and is not the overall retrieval estimate; Table~\ref{tab:6} examines the four negative passage-use cases.}
\label{fig:8}
\end{figure*}

Figure~\ref{fig:8} presents the primary full-cohort outcome account alongside a secondary sensitivity display: Figure 8B gives the outcome classification for all 74 pairs, whereas Figure 8A gives the conditional 70-pair mean comparison. The full 74-pair outcome classification comprises 11 improved, 52 neutral, four negative, and seven unassessable pairs. Negative effects range from an unnecessary digression to displacement of the requested action; the label describes answer degradation linked to returned content, not observed patient injury. The full-cohort paired estimates reported below remain the primary comparison.

For all 74 matched questions, the paired mean differences (RAG-on minus RAG-off) were usefulness -0.041 (95\% CI -0.243 to +0.149); safety +0.014 (95\% CI -0.135 to +0.149); correctness +0.014 (95\% CI -0.135 to +0.162); actionability 0.000 (95\% CI -0.203 to +0.189). Each interval included zero. The intervals quantify uncertainty in the recorded ratings; they do not remove the value of case-specific improvements or the need to investigate negative changes.

For the secondary 70-pair sensitivity comparison, the corresponding differences were usefulness +0.029 (95\% CI -0.157 to +0.200); safety +0.086 (95\% CI -0.043 to +0.214); correctness +0.057 (95\% CI -0.100 to +0.214); actionability +0.057 (95\% CI -0.129 to +0.243). All four intervals also included zero. Because the four pairs were excluded after outcome review, these favorable conditional differences cannot be interpreted as an unbiased estimate or as evidence that retrieval was superior. They are informative only as a sensitivity decomposition: they show the direction among the non-flagged responses while the primary 74-pair analysis retains every observed outcome. This reporting makes the exclusion transparent rather than allowing it to function as selective reassurance.

The outcome proportions are 11/74 improved (14.9\%), 52/74 neutral (70.3\%), 4/74 negative (5.4\%), and 7/74 unassessable (9.5\%). The corresponding 95\% Wilson intervals are 8.5--24.7\%, 59.1--79.5\%, 2.1--13.1\%, and 4.7--18.3\%, respectively. The four negative cases appear in Table~\ref{tab:6}; none represents an observed patient injury.

\subsection{Response length and retrieval overhead}
All 74 paired runs completed without recorded runtime failures or empty answers. Mean output length was 102.39 tokens with RAG-on and 123.89 with RAG-off. Mean first-token delay was 6.12 versus 5.52 seconds, and mean generation time was 10.83 versus 11.52 seconds. Local retrieval took a mean of 1.069 seconds and was timed separately from generation. Mean sampled process memory was similar (2.46 versus 2.47 GB) on the 6 GB Android CPU emulator. These averages describe response usage and execution for the paired workload; generation time alone excludes retrieval overhead.

\subsection{Failure mechanisms and corrective targets}
Mean scores summarize average rubric ratings across paired responses. The case audit then examines each question, its reviewed action, the returned passage, and the generated answer to explain why a pair improved, remained neutral, became negative, or could not be attributed to retrieval. This keeps the response-level mechanism visible alongside the aggregate comparison and follows behavioral evaluation approaches that test a defined capability under its relevant conditions \cite{r4,r27,r31}.

Table~\ref{tab:6} examines four passage-linked negative cases. The returned passage, its fit to the requested action, and the observed response change are shown separately; seven lower-rated pairs without established passage causation are discussed afterward.

\begin{table*}[!t]
\centering
\caption{Negative passage-linked effects: returned content, fit, and response effect}
\label{tab:6}
\footnotesize
\setlength{\tabcolsep}{5pt}
\renewcommand{\arraystretch}{1.12}
\begin{tabularx}{\textwidth}{@{}p{0.18\textwidth}p{0.25\textwidth}p{0.19\textwidth}X@{}}
\toprule
\textbf{Clinical request} & \textbf{Returned passage} & \textbf{Fit to request} & \textbf{Observed answer; corrective target} \\
\midrule
Headache with visual changes & Weight-gain guidance mentioning urgency only alongside rapid weight change, swelling, or severe headache. & Outside the immediate referral and handover request. & RAG-on imported body-mass-index and weight-trend details that were not requested. Corrective target: keep the response focused on the reported warning signs and the requested handover. \\
Rapid weight gain, swelling, headache & Cholestasis guidance about nocturnal itching and reduced fetal movement. & Irrelevant to the symptoms presented. & RAG-on retained urgent clinic contact but added itching advice from the passage. Reject unrelated passages to protect answer focus. \\
Maternal/newborn records & Pregnancy and breastfeeding nutrition guidance. & Irrelevant to which records to keep. & RAG-on answered about diet instead of records. Reject action/topic mismatches before passage use. \\
Routine infant vaccination & Breastfeeding compatibility, vaccine scheduling, monitoring and medicine cautions. & Partially relevant to vaccination but does not answer why vaccines are needed. & RAG-on gave the core benefit but added breastfeeding, reaction and medicine advice; focus on vaccine protection was reduced. Limit evidence use to the requested explanation. \\
\bottomrule
\end{tabularx}
\end{table*}

The seven unassessable pairs have at least one lower-rated aspect, but the trace does not establish that retrieval caused the difference. The category is heterogeneous. In the nocturnal-itching case, the passage correctly calls for prompt contact, but RAG-on softens that action to mentioning it at the next visit; the defect is application of suitable evidence, not a false source claim \cite{r57}. The blood-pressure and antenatal-start cases contain related material, but the model does not answer the specific question as directly as RAG-off. The postnatal passage is only a question heading, so its contribution to the answer cannot be determined. In the nausea-medication case, the retrieved material is poorly aligned and includes self-care advice, yet the model resists that direction and refuses to endorse an unidentified tablet. Likewise, the warning-sign passage in the induction case and the diabetes passage in the convulsion case do not enter the RAG-on answers. These examples show that the model can sometimes ignore or contain unsuitable evidence, even when the resulting answer remains less useful than RAG-off. Unassessable therefore includes both uncertain passage effects and successful mitigation of unsuitable context; it does not mean that the passage was useful or that the answer was fully satisfactory \cite{r31,r59,r60,r61}.

The unassessable cases therefore include two kinds of behavior: the model sometimes fails to apply a fitting passage precisely, and sometimes limits the influence of a poorly matched passage. The latter is visible in the unknown-tablet question, where the model refuses to recommend medication despite the retrieved self-care material, and in the convulsion question, where irrelevant diabetes content is not repeated. Such resistance is consistent with retrieval-robustness work showing that generators may handle distracting context unevenly \cite{r59,r61}. The neutral category is narrower: it denotes a response that is adequate and comparable to RAG-off without a material passage-linked gain or loss. It does not require every retrieved sentence to appear in the answer.

The four negative cases instead show returned material entering the answer and reducing focus or appropriateness. Three passages were off topic for the requested action: weight-gain information entered a headache handover, itching advice entered a swelling-and-headache response, and nutrition replaced a maternal-and-newborn-records answer. In the vaccination case, a partly related passage about breastfeeding and post-visit monitoring added tangential advice to a question asking why vaccines matter. The role of irrelevant and superficially related passages as distractors is supported by controlled RAG studies; the four case classifications themselves derive from the preserved traces in this experiment \cite{r28,r59,r60}.

The corrective targets include stronger alignment of retrieved text with the current question and a generation instruction to use only evidence needed for that question while preserving required actions. Filtering unhelpful snippets has improved medical RAG in other settings, and robustness studies show that the generator's treatment of distractors matters as well \cite{r60,r61}. A revised instruction is a plausible mitigation for overuse in this system; its effect was not measured in the present experiment.

\section{Discussion}
\subsection{Principal findings and the quality--resource trade-off}
The candidate comparison selected Gemma because it better maintained clinical context across the scored conversations, despite Qwen's lower memory use and faster responses. Clinician scoring then found that 147 of 150 isolated Gemma answers met the acceptance criterion. The retrieval experiment gave a less uniform result: 11 of 74 paired answers improved, 52 were comparable, four used returned content in a way that degraded the answer, and seven were lower-rated without clear passage causation. The phone test showed completed offline generation on one configuration. These results answer different parts of the system question and should not be combined into a single success rate.

The competing candidates expose a concrete selection trade-off. Qwen generated faster and used less process memory, but its weaker retention of earlier facts made it less suitable for a chatbot whose later turns depend on what the person already said. The staged evaluation therefore took Gemma into the 150-question clinician review and the retrieval study. Because each artifact ran in its native runtime, the measured differences describe complete application configurations. Mobile inference research likewise treats quality and execution cost as joint design considerations \cite{r13,r37,r38}.

The clinician ratings reveal a different issue from context loss. Correctness and clarity were high, while completeness was lower: some answers were accurate and readable yet omitted part of the requested guidance. The clinical key defines which details matter for that question. It also keeps the paired judging from rewarding a long answer that repeats background information while missing the required action \cite{r4,r5,r11}.

\subsection{Device envelope and offline application operation}
Physical-device testing places the model comparison in the context of an integrated Android application. All nine recorded offline requests completed, with a median of 8.6 s before the first token and 14.3 s for the full response. The difference between those measures matters because the user waits for an answer to begin and then for the requested guidance to finish. This result demonstrates a practical local response path in the tested configuration, while the emulator measurements remain useful for comparing candidate runtimes under their controlled workloads \cite{r37,r38}.

Offline operation also depends on what happens when connectivity returns. The application generated answers locally during internet loss and retained conversation and telemetry records for later acknowledgement by the connected service. In the integration record, 46 of 47 synchronization measurements succeeded. This makes local answering and deferred data transfer parts of one application workflow: the answer does not need a live connection, and pending records can subsequently reach the service. Such continuity is relevant to maternal-health information access where connectivity varies, consistent with digital-health implementation guidance \cite{r18,r19}.

The handset profile adds a useful distinction to the performance evaluation. Median generation throughput of 22.2 tokens/s describes output once generation is under way, whereas the 8.6 s first-token delay and 14.3 s total latency describe the wait experienced during the recorded requests. A single throughput figure would conceal that initial delay. The nine requests did not invoke retrieval, so the physical-device timing establishes a local answer baseline for this workload; the clinical question-bank and paired retrieval experiments supply separate evidence about response quality and evidence use \cite{r13,r37,r38}.

\subsection{Why validated knowledge can still degrade an answer}
The primary 74-pair analysis is the main retrieval comparison. It retains every matched question, reports the full paired differences and uncertainty intervals, and accounts for all 11 improved, 52 neutral, four negative, and seven unassessable outcomes. The 70-pair display is a secondary sensitivity analysis. It removes the four audit-flagged pairs only to show whether the descriptive direction among the remaining responses is similar when the mechanism-level cases are examined separately. Because this removal followed outcome review, the display is not an unbiased effect estimate and cannot be used to claim overall retrieval superiority. Its value is interpretive: it decomposes the ordinary paired responses while the complete 74-pair analysis remains the basis for the overall result \cite{r4,r6,r27,r31}.

The paired experiment separates clinical content validation from evidence use. A passage can be valid for its original question and still be inappropriate for the current life stage, urgency, or practical request. An answer can also repeat a correct source statement while omitting the action required by the current symptoms. The record-keeping and escalation failures demonstrate that the relevant unit of evaluation is the generated response to the complete question, rather than the provenance of an isolated passage. Medical RAG work similarly distinguishes helpful context selection from the quality of the eventual answer \cite{r31,r60}.

The findings connect with medical RAG research in which component choices affect performance \cite{r3,r60}, and with controlled evidence that irrelevant or superficially related passages can distract a generator \cite{r28,r59}. In the records case, nutrition guidance displaced the record-keeping request. In the headache case, weight-gain guidance entered the handover details even though the request concerned immediate referral and a minimal handover. A third off-topic passage about itching entered an otherwise urgent response to rapid weight gain and headache, showing direct contamination of the answer by irrelevant evidence without changing the urgent-care action. These are distinct degrees of negative passage use. The vaccination case illustrates a subtler focus loss: a partly relevant passage contributed extra details without overturning the central answer. The itching question itself is different: its relevant passage called for prompt assessment, but the answer softened that instruction \cite{r57}. The dating answer introduced an hCG claim absent from its passage. Reviewing the exact returned text prevents a generated error from being automatically attributed to the clinically validated package.

Local FTS5/BM25 ranking uses lexical overlap and explicit admission checks rather than semantic embeddings \cite{r29}. This keeps retrieval available on device without installing a second model or adding its memory and compute costs, an important design trade-off for offline execution \cite{r13,r37}. It can nevertheless admit passages sharing maternal-health terms without matching the user's requested action. The weight-gain, nutrition, and unrelated itching matches illustrate this weakness in the tested corpus. The relevant cholestasis passage shows why changing retrieval alone would not resolve every failure: the generator must apply the clinical action in the passage. Research on distracting passages and medical snippet filtering supports assessing both passage fit and answer behavior, while the feasibility of any alternative retriever on this handset remains untested \cite{r31,r59,r60}.

Consumer-health RAG studies likewise find that retrieval benefits depend on the question and evidence used \cite{r35}. In this cohort, the requested care action proved more discriminating than broad topical overlap.

The case audit defines concrete acceptance criteria for a retrieval iteration. Higher coverage is desirable only when the additional passages maintain or improve the final response. A retrieval change should be evaluated against the full question, including danger signs and requested actions, and against a separate set of newly reviewed questions. Rerunning only the identified failures would demonstrate repair of those examples but would not establish general improvement. The full 74-pair account remains necessary because removing selected flagged cases changes the question answered by the mean comparison.

\subsection{Selective grounding in maternal-health conversation}
The paired findings provide a practical rationale for selective evidence acquisition. Most pairs were neutral, 11 had passage-supported improvements, four had negative passage-linked effects, and seven had lower-rated aspects with uncertain passage attribution. These outcomes argue for deciding when evidence is useful and checking how it influences the answer, rather than equating additional context with better assistance. The current model-initiated design makes that choice explicit through a bounded local tool. Its value for conversation is the ability to move between a direct explanation and a documented lookup as the information need changes, without making retrieval an unavoidable part of every exchange \cite{r49,r50,r51}.

The safety boundary continues across both paths. A direct response must respect uncertainty and urgency; a retrieved response must additionally establish that the supplied material fits the full question. The four flagged response traces illustrate why invocation and appropriate evidence application require separate assessment, consistent with medical-RAG and context-quality evaluation \cite{r3,r31}. Local source restriction, deterministic admission, and the instruction to ignore mismatched evidence provide complementary controls. The literature on retrieval decisions likewise shows that confidence and knowledge sufficiency are not interchangeable \cite{r52}.

The distinction between invocation, admission, and application also improves interpretability of reliability. Tool events show whether evidence was requested, passage records show what was returned, and response assessment shows whether the required facts and action survived generation. A no-retrieval answer is appropriate when the information need is met safely; a tool call is useful only when it contributes suitable evidence. This process-level interpretation follows agentic-RAG evaluation concerns while retaining the clinical and conversational criteria of the present study \cite{r53,r4,r31}.

\subsection{Relationship to maternal health research and deployment}
HerCare and the maternal chatbot work in India emphasize retrieval-supported answers and expert assessment \cite{r2,r7}. MamaBench tests responses to clinically decisive changes in a presentation \cite{r26}. Here, the same concern appears across a sustained exchange and when retrieved evidence enters the prompt: the answer must still address the person's current request and required action.

ADVISER demonstrates how field constraints can alter the implementation of an AI intervention \cite{r25}. For an offline assistant, the corresponding operational questions include which devices can sustain the workload, how users encounter and understand escalation guidance, whether retrieval and response behavior remain appropriate, and how knowledge is updated. These operational considerations connect system evaluation to implementation planning, as emphasized in digital-health reporting and intervention guidance \cite{r18,r19}.

The local execution goal also has a distributional consequence. Removing a network dependency can improve availability for a supported device while leaving lower-resource devices unable to run the complete configuration. The physical-device findings in Section~\ref{sec:physical-device} make this visible. Claims of accessibility should therefore specify the installed artifact size, usable memory, workload, and device conditions, rather than assuming that offline availability implies suitability for every entry-level handset. The access motivation and the operating configuration are therefore treated together in deployment planning \cite{r18,r23}.

The handset result supports feasibility for the tested application configuration. Local responses completed under the recorded workload, and pending records were acknowledged after reconnection. Prospective clinical supervision, user testing, and a wider device sample are the next stages for evaluating use in care settings, following DECIDE-AI and mHealth reporting guidance \cite{r20,r19}.

\subsection{Implications for evaluation and continued development}
The question keys make scores traceable to a missing fact, an incorrect care action, lost conversation context, or an observed tool call. A single preference score would obscure those differences. HealthBench provides a precedent for case-specific clinical criteria; G-Eval and Prometheus show how explicit rubrics can guide structured review \cite{r4,r30,r32}. That traceability helps identify which behavior a later system change should improve.

The distinction between relevance, groundedness, and correctness is especially useful for retrieval development. A passage can be relevant without supporting every generated claim, and an answer can remain correct despite ignoring a poor passage. RAGAs formalizes these as separate dimensions of pipeline evaluation \cite{r31}. The present traces add the requested clinical action to that distinction, allowing a retrieval revision to be checked for both evidential fit and preservation of urgency.

Continued development can build on these findings through targeted tests of escalation, requested-action alignment, conversation context, and evidence use. The benchmark action interface remains a separate instrument for testing structured tool decisions and argument validation. New knowledge-package versions should be assessed against the complete response rubric and the retained failure cases, with additional questions testing the same mechanisms. CheckList supports this capability-oriented approach to regression testing \cite{r27}. Further system evaluation can be reported using the applicable TRIPOD-LLM and mERA principles \cite{r6,r19}. For subsequent clinician-supervised evaluation, DECIDE-AI provides a framework for reporting early-stage clinical studies of AI decision-support systems \cite{r20}.

\subsection{Limitations}
The knowledge package was curated within a single organization, and evaluation covered two compact configurations with different native runtimes. Findings therefore concern the adopted configurations and the studied maternal-health question distribution. The fixed benchmark and simulated conversations do not establish live-user behavior or clinical outcome effects. Physical-device inference and telemetry synchronization were observed through the integrated application, but these observations do not establish performance across Android devices.

The nine handset inference requests did not measure retrieval decisions or clinical response quality. Those outcomes were assessed in separate benchmark workloads. The handset and synchronization observations establish that the application completed these functions under the recorded conditions; they cannot establish sustained user acceptance or clinical reliability in daily use.

Retrieval was accepted for 74 of 150 questions, so the primary paired response comparison describes the retrieval-eligible subset. Small outcome groups, ordinal ratings, and an on-then-off collection order limit the precision and generality of the descriptive differences. The complete 74-pair estimates and outcome panel retain every pair. The conditional 70-pair means are a secondary, outcome-selected sensitivity analysis after the four audit-flagged pairs were set aside for their case-level mechanisms; they are not an unbiased overall effect estimate and are not used to replace the complete analysis. RAG-on changed both passage availability and an evidence-use instruction, so their separate contributions cannot be estimated from these pairs. The four negative labels identify visible uptake of distracting material, while the seven unassessable labels retain lower-rated answers without a supported passage attribution. A different prompt and an embedding-based retriever were not tested as remedies for these mechanisms, so their ability to reduce the observed focus and evidence-application problems remains unknown \cite{r6,r59,r60,r61}.

LLM-based judgments of context retention can vary across repeated assessments and with changes in evaluation wording, even when explicit rubrics guide scoring \cite{r54,r55}. The context-retention results should therefore be read as rubric-based conversational evidence. Clinically reviewed safety ratings remain in the single-question assessment, while the scripted tool score supplies the application-action evidence.

\subsection{Future outlook and post-implementation evaluation}
CPU inference optimization should examine operator and kernel efficiency, thread scheduling, prompt processing, key--value cache management, context budgeting, and cold-start cost. Controlled comparisons of inference algorithms and runtime configurations should determine which changes improve latency and request completion on constrained Android devices, following the joint quality and resource assessment used in mobile inference research \cite{r37,r38}.

Further physical-device evaluation should repeat matched selected-model workloads across 6 GB, 8 GB, and 12 GB RAM classes. Loading, steady-state and peak memory, initialization, response latency, throughput, completion, and failure measurements should be assessed alongside battery and thermal behavior. This would test whether additional memory and compute translate into a more dependable interaction envelope while preserving response quality \cite{r37,r38}.

Future evaluation can repeat the context-retention judgments with order-reversed runs and compare the scripted tool assessment across repeated executions \cite{r11}.

Real-world pilot evaluation should assess adoption, usability, task completion, device and network reliability, synchronization, clinical safety, user experience, staff workload, operational cost, and implementation barriers. A structured protocol should define participant criteria, ethics approval and consent, clinical oversight, and safety reporting before recruitment. DECIDE-AI provides guidance for early clinical evaluation, with SPIRIT-AI applicable to any subsequent interventional trial protocol \cite{r20,r41}.

\section{Conclusion}
Gemma was selected for the maternal-health assistant because it retained context better than Qwen in the conversation benchmark; Qwen was faster and used less memory. Clinicians accepted 147 of 150 Gemma answers to the fixed question bank. Supplying local passages improved 11 of 74 matched answers, left 52 comparable, and degraded four through identifiable passage use; seven lower-rated pairs had uncertain passage attribution. The full-cohort mean differences were small and uncertain. All nine recorded offline requests completed on the tested 6 GB handset, with a median total response time of 14.3 s, and the application synchronized pending records after reconnection. The four degraded answers, particularly where returned material displaced the requested care action, make passage fit and preservation of clinical urgency central safety checks for the next evaluation. Together, the findings characterize the integrated offline maternal-health assistant within the designed and tested benchmark, retrieval pipeline, and handset conditions.

\section*{Data and Code Availability}
The implementation and a minimal reproducibility package are currently maintained in a private project repository and are available only upon request. The package contains the LaTeX sources, figure sources, table inputs, frozen non-sensitive benchmark and paired-analysis outputs, summary-statistics and plotting scripts, model and runtime manifests, and a README describing the software environment and regeneration steps. These materials are intended to regenerate the reported tables, figures, and derived summary values without rerunning the model. Clinical question text, response-level clinical review records, and other restricted source materials remain governed separately; where those materials cannot be released, the package provides the corresponding frozen aggregates, identifiers, and non-sensitive metadata needed to inspect the reported calculations. This structure follows reproducible reporting principles for LLM studies while preserving the access controls applied to clinical review materials \cite{r6,r19}.

\appendices
\section{Paired Cohort Accounting}
Figure~\ref{fig:8} contains the primary full-cohort outcome account and a secondary sensitivity display. Figure 8B reports the outcome categories for the complete 74-pair cohort. Figure 8A reports mean rubric scores for the 70 matched questions retained after the same four audit-flagged pairs were excluded from both conditions. The plotted means and the primary complete-cohort estimates are described in Section~\ref{sec:rag-results}; the complete analysis retains all 74 questions. Table~\ref{tab:6} details the four negative passage-linked cases, including two within the 70-pair sensitivity subset, while the seven unassessable cases are explained in the results text. The excluded identifiers are T05-PREG-002, T05-ANC-020, T05-COM-009, and T05-POST-020; exclusion applies to both response rows for each identifier.

The retrieval-effect counts sum to all 74 pairs: 11 improved, 52 neutral, 4 negative, and 7 unassessable. These annotations are retained independently of the paired rubric scores. Numeric ratings remain possible for the seven pairs with unassessable retrieval attribution. The source texts, clinical keys, and retrieved passages remain the evidence records for the paired analysis; the outcome labels and scores are reported as the current analysis of those records.

\section{Reproducibility and Additional Execution Evidence}
\subsection{Minimal reproducibility package}
The reproducibility package is organized around frozen outputs rather than access to the original clinical conversation export. It preserves the input records used for reported aggregate values, the scripts that derive table and figure data, figure-source files, artifact and runtime manifests, and the environment and command notes required to regenerate the manuscript outputs. Regeneration from this package reproduces the reported descriptive statistics, confidence intervals, tables, and figures from frozen outputs; this is a derived-analysis reproducibility claim, not a claim that the original model-generation or LLM-judge execution can be rerun. The question bank, response-level review packet, and other restricted records remain available only through the applicable access process.

\subsection{Generation and retrieval settings}
Both candidate configurations used CPU inference. The 150-question clinical response collection used an approximately 8 GB Android emulator and followed its own fixed response-generation procedure. The later paired evidence-context study used a separate 6 GB emulator and is described independently in Section~\ref{sec:rag-results}; its retrieval-eligibility count belongs only to that paired comparison, not to the clinician-reviewed response collection.

Conversation assembly used a 4096-token runtime ceiling, at most five completed user--assistant turns, and a 3072-token history budget within the total prompt limit. Retrieval used SQLite FTS5 with BM25 ranking. Raw confidence of at least 95 entered the high-confidence pool; scores from 65 to below 95 required metadata and action checks. Pool scores were normalized and passages below 80 were removed, with a maximum of three retained passages. These are implementation thresholds, not probabilities of clinical relevance. The current model-facing retrieval tool uses the same bounded search service but returns no more than two passages per invocation, with at most two tool-call rounds. The three-passage limit describes the underlying candidate pool and the recorded evidence-context collections; it is not the current tool-output limit.

\begin{table}[!tbp]
\centering
\caption{Adopted artifact and inference configuration for reproduction. Artifact integrity hashes are retained in the repository specification.}
\label{tab:7}
\small
\begin{tabular}{@{}>{\raggedright\arraybackslash}p{.23\linewidth}>{\raggedright\arraybackslash}p{.35\linewidth}>{\raggedright\arraybackslash}p{.32\linewidth}@{}}
\toprule
\textbf{Configuration} & \textbf{Artifact and precision} & \textbf{Runtime and controls} \\
\midrule
Gemma 4 E2B & \texttt{gemma-4-E2B-it.}\newline\texttt{litertlm}; mobile mixed 2/4/8-bit scheme & LiteRT-LM Android 0.18.0; native tokenizer/template; CPU \\
Qwen3.5 0.8B & \texttt{Qwen3.5-0.8B-}\newline\texttt{Q4\_K\_M.gguf}; four-bit class & Android llama.cpp; GGUF tokenizer/template; CPU \\
Paired RAG run & Same selected Gemma artifact and 21,673-passage package in both arms & 6 GB Android guest; temperature 0.0; 512 output tokens; seed 17; tools disabled \\
\bottomrule
\end{tabular}
\end{table}

\subsection{Selection indices and timing definitions}
The conversation index is 100 times the equal mean of context retention/4 and (benchmark tool behavior + 5)/10. The resource index is defined as follows. Let $G=8{,}129{,}460$~kB be the observed capacity of the 8~GB reference guest, $R_i$ the recorded peak process RSS for configuration $i$ in the same units, $T_i$ its mean conversation duration in seconds, and $N_i$ its clean-turn count out of $N=312$. Memory headroom, relative speed, clean completion, and the resource index are
\begin{align}
M_i &= 100\frac{G-R_i}{G},\\
S_i &= 100\frac{\min_j T_j}{T_i},\\
C_i &= 100\frac{N_i}{N},\\
I_{\mathrm{resource},i} &= \frac{M_i+S_i+C_i}{3}.
\end{align}
Thus, $S_i$ uses mean conversation duration; isolated-response latency, throughput, and first-token delay are reported separately and do not enter this index. Using the recorded values gives $(M,S,C)=(48.0,46.8,95.2)$ and $I_{\mathrm{resource}}=63.3$ for Gemma, and $(79.2,100.0,100.0)$ and $I_{\mathrm{resource}}=93.1$ for Qwen. Figure~\ref{fig:5} shades the conversation and resource indices using the same project decision illustration. These drawing rules are not clinical acceptance boundaries.

The paired timing record times retrieval separately from generation; its total-latency label repeats recorded generation time. It should not be interpreted as a directly measured sum of all interaction stages. First-token values are reported only when the relevant runtime instrumentation supplies them. Peak RSS is a sampled process maximum, not a continuous trace of device memory.

\subsection{Clinician-rating agreement estimator}
Absolute agreement among the three retained rating columns uses the two-way single-rating estimator \cite{r47}
\begin{equation}
\begin{aligned}
\mathrm{ICC}(A,1)&=\frac{MS_Q-MS_E}{D},\\
D&=MS_Q+(k-1)MS_E\\
 &\quad+\frac{k(MS_R-MS_E)}{n}.
\end{aligned}
\label{eq:icc}
\end{equation}
Here $MS_Q$, $MS_R$ and $MS_E$ are the question, reviewer and residual mean squares, respectively; $n=147$ complete questions and $k=3$ rating columns. Question-level resampling with 2,000 replicates supplies the percentile intervals. This single-rating coefficient describes agreement within the retained reviewer panel; it is algebraically equivalent to the ICC(2,1) coefficient in the Shrout--Fleiss notation \cite{r47,r48}.

\section{Rating Definitions and Interpretation}
For the four displayed RAG dimensions, the working rubric uses 0 for an unusable response, 1 for a major failure, 2 for partial performance, 3 for a useful response with a limitation, and 4 for a sufficient and appropriate answer to the actual question. Usefulness does not require exhaustive optional content. Correctness concerns claims made; omissions are evaluated through the relevant usefulness or essential-content requirement. Safety includes urgency for current danger signs. Actionability includes a usable explanation when that is what the user requested.

Completeness, clarity, directness, groundedness, and evidence-use fields retain earlier rating passes and were not substituted for the revised displayed measures. Groundedness is inapplicable to RAG-off. Critical-error and warning-sign fields are separate from the four retrieval-audit flags used for the conditional display. The working workbook preserves these distinctions and question-specific notes.

\end{document}